\documentclass[acmtog, authorversion, nonacm]{acmart}

\usepackage[utf8]{inputenc}

\usepackage{amsthm}
\usepackage{amsfonts} 
\usepackage{mathtools} 
\usepackage{siunitx}

\usepackage{ifthen}
\usepackage{cleveref}
\usepackage{calc}

\usepackage{algorithm}

\usepackage{multirow}

\usepackage{subcaption}
\usepackage{verbatim}
\usepackage{xspace}

\DeclareMathOperator*{\argmin}{arg\,min}
\usepackage{amsfonts}

\definecolor{GiancarloColor}{rgb}{0,0.70,0.20}
\definecolor{YidanColor}{rgb}{0,0.20,0.70}
\definecolor{YuriiColor}{rgb}{0.54,0.1,0.25}
\definecolor{DanieleColor}{rgb}{0.56,0.34,0.62}
\definecolor{DavidColor}{rgb}{0.56,0,1}
\definecolor{ClaudioColor}{rgb}{0.0,0.,1.}

\newcommand{\DP}[1]{{\leavevmode\color{DanieleColor} Daniele: #1}}
\newcommand{\GP}[1]{{\leavevmode\color{GiancarloColor} Giancarlo: #1}}
\newcommand{\YG}[1]{{\leavevmode\color{YidanColor} Yidan: #1}}
\newcommand{\YP}[1]{{\leavevmode\color{YuriiColor} Yurii: #1}}
\newcommand{\CS}[1]{{\leavevmode\color{ClaudioColor} Claudio: #1}}
\newcommand{\DF}[1]{{\leavevmode\color{DavidColor} David: #1}}
\newcommand{\systemname}[0]{\textit{LookUp3D}\xspace}

\newcommand{\rev}[1]{{\color{blue}{#1}}}

\newcommand{\nothing}[1]{}

\providecommand{\finalversion}{1} 
\ifthenelse{\equal{\finalversion}{1}}
{
	\renewcommand{\DP}[1]{}
	\renewcommand{\GP}[1]{}
	\renewcommand{\YG}[1]{}
	\renewcommand{\YP}[1]{}
	\renewcommand{\DF}[1]{}
	\renewcommand{\CS}[1]{}
	\renewcommand{\rev}[1]{{\color{black}{#1}}}

}

\definecolor{forestgreen}{rgb}{0.13,0.54,0.13}
\definecolor{darkblue}{rgb}{0,0,.5}
\hypersetup{
	unicode=true,
	colorlinks=true,
	citecolor=forestgreen, 
	linkcolor=darkblue, 
	urlcolor=darkblue, 
}

\crefname{appsec}{Appendix}{Appendices}

\let\originalleft\left
\let\originalright\right
\renewcommand{\left}{\mathopen{}\mathclose\bgroup\originalleft}
\renewcommand{\right}{\aftergroup\egroup\originalright}

\DeclareFontFamily{U}{mathx}{\hyphenchar\font45}
\DeclareFontShape{U}{mathx}{m}{n}{<-> mathx10}{}
\DeclareSymbolFont{mathx}{U}{mathx}{m}{n}

\usepackage{shortbold}
\graphicspath{{images/}}

\acmSubmissionID{2209}

\setcopyright{acmlicensed}
\acmConference[SA Conference Papers '25]{SIGGRAPH Asia
2025 Conference Papers}{December 15--18, 2025}{Hong Kong, Hong Kong}
\acmBooktitle{SIGGRAPH Asia 2025 Conference Papers (SA Conference Papers '25),
December 15--18, 2025, Hong Kong, Hong Kong}
\acmDOI{10.1145/3757377.3763986}
\acmISBN{979-8-4007-2137-3/2025/12}

\begin{document}
\title{\systemname: Data-Driven 3D Scanning}

\author{Giancarlo Pereira}
\authornote{Joint first authors with equal contribution.}
\orcid{0000-0002-5869-6886}
\email{giancarlo.pereira@nyu.edu}
\affiliation{%
    \department{Tandon School of Engineering}\institution{New York University}
    \city{New York City}
    \country{USA}    
}

\author{Yidan Gao}
\authornotemark[1]
\orcid{0009-0005-0321-0470}
\email{yidan.gao@nyu.edu}
\affiliation{%
\department{Courant Institute of Mathematical Sciences}\institution{New York University}
        \city{New York City}
    \country{USA}    
}

\author{Yurii Piadyk}
\authornotemark[1]
\orcid{0000-0002-2975-635X}
\email{ypiadyk@nyu.edu}
\affiliation{%
    \department{Tandon School of Engineering}\institution{New York University}
        \city{New York City}
    \country{USA}    
}

\author{David Fouhey}
\orcid{0000-0001-5028-5161}
\email{david.fouhey@nyu.edu}
\affiliation{%
    \department{Courant Institute of Mathematical Sciences, Tandon School of Engineering}\institution{New York University}
        \city{New York City}
    \country{USA}    
}

\author{Claudio T. Silva}
\orcid{0000-0003-2452-2295}
\email{csilva@nyu.edu}
\affiliation{%
    \department{Center for Data Science, Tandon School of Engineering}\institution{New York University}
        \city{New York City}
    \country{USA}    
}

\author{Daniele Panozzo}
\orcid{0000-0003-1183-2454}
\email{panozzo@nyu.edu}
\affiliation{%
    \department{Courant Institute of Mathematical Sciences}\institution{New York University}
        \city{New York City}
    \country{USA}    
}





\begin{abstract}
High speed, high-resolution, and accurate 3D scanning would open doors to many new applications in graphics, robotics, science, and medicine by enabling the accurate scanning of deformable objects during interactions. 
%
Past attempts to use structured light, time-of-flight, and stereo in high-speed settings have usually required tradeoffs in resolution or inaccuracy.
%
In this paper, we introduce a method that enables, for the first time, 3D scanning at 450 frames per second at 1~Megapixel, or 1,450  frames per second at 0.4~Megapixel in an environment with controlled lighting.
%
The key idea is to use a per-pixel lookup table that maps colors to depths, which is built using a linear stage. Imperfections, such as lens-distortion and sensor defects are baked into the calibration.
%
We describe our method and test it on a novel hardware prototype. We compare the system with both ground-truth geometry as well as commercially available dynamic sensors like the Microsoft Kinect and Intel Realsense. Our results show the system acquiring geometry of objects undergoing high-speed deformations and oscillations and demonstrate the ability to recover physical properties from the reconstructions.
\end{abstract}

%
%

\ccsdesc[500]{Computing methodologies~3D imaging}
\ccsdesc[500]{Computing methodologies~Active vision}
\ccsdesc[500]{Computing methodologies~Computational photography}
\ccsdesc[500]{Computing methodologies~Reconstruction}
\ccsdesc[300]{Hardware~Emerging optical and photonic technologies}
\ccsdesc[300]{Hardware~Scanners}
%
%

\keywords{3D Scanning, Geometry Acquisition,  Structured Light, Data-Driven, Active Illumination, High-Speed}

\begin{teaserfigure}
    \centering
        \includegraphics[trim={500pt 0 300pt 0}, clip, width=\linewidth]{fig/teaser/teaser_v2.pdf}
    \caption{
        \textbf{Reconstructing Dynamic Motion at 450~fps.} We visualize point cloud reconstructions of a silicone bunny dropping, with consecutive frames being 22 milliseconds apart. \systemname reconstructs dense 3D geometry at each frame, accurately capturing both deformation and motion despite moderate blur and 1-millisecond exposure. We further validate reconstruction fidelity by estimating gravitational acceleration at 9.60\,m/s\textsuperscript{2} (versus its ground truth value of 9.80\,m/s\textsuperscript{2}) from the recovered trajectories, demonstrating the utility of our method for downstream tasks such as physical inference and inverse design.
    }
    \Description[Teaser Figure]{Point clouds of a bunny in free-fall, showing capability of LookUp3D to capture motion at high-speeds.}
    \label{fig:teaser}
\end{teaserfigure}

\maketitle

\section{Introduction}
\label{sec:introduction}


3D scanning is ubiquitous in modern technology, enabling industrial metrology, face recognition, cultural heritage preservation, and immersive VR. Various solutions have been introduced to tackle this problem, including touch probes \cite{dobosz_cmm_2005}, time-of-flight \cite{tang_automatic_2010}, 
structured light \cite{salvi_pattern_2004}, polarized light \cite{deep_polarization_2021}, and event cameras \cite{EMVS_2018}. Among these, structured light (SL) stands out as a non-contact, high-resolution technique that delivers consistent, repeatable results across diverse materials and object sizes. 
SL systems typically reconstruct 3D geometry using a calibrated projector-camera pair to project time-varying patterns (e.g., phase-shifted sinusoids or Gray Codes) and analyze their deformations. Through epipolar constraints and triangulation, pixel correspondences are established, achieving submillimeter precision.
%
Despite six decades of advancements in coding schemes \cite{mirdehghan_optimal_2018, adaptive_color_structured_light}, hardware synchronization \cite{zhang_superfast_2010, Koppal-2012-120215}, and hybrid sensing \cite{event_camera_sl_2021, event_camera_sl_2023}, two fundamental challenges persist: (1) SL accuracy is bounded by triangulation, which in turn is bounded by the calibration quality of the projector-camera setup; (2) SL capture speed is constrained by the need to project and decode multiple patterns per scan to compute correspondences for triangulation.


In this work, we introduce \systemname, a novel SL scanning approach that eliminates the need for triangulating correspondences in a calibrated camera-projector system. This shift enables reconstruction with fewer projected patterns and supports higher-speed acquisition for dynamic scenes. \systemname treats the projector as a simple light source rather than a generator of precise codes, operating with as few as one RGB pattern, and performs depth reconstruction through a per-ray lookup table (LUT) obtained by actual recordings during calibration. By avoiding triangulation between projector and camera, our method eliminates the need to model projector intrinsics or align projector-camera geometry precisely -- they all get implicitly baked into the LUT. This allows the system to tolerate significant optical distortions or misalignment on the projection side. The recorded LUT not only enables direct, highly parallelizable depth retrieval, but also provides a measure of reconstruction confidence. Each pixel’s match quality -- how closely the observed color aligns with the calibration data -- serves as an indicator of reliability, allowing for automatic filtering of poorly reconstructed regions.
%
%
We validate our approach across three hardware configurations: a standard DLP projector, an inexpensive LCD projector, and a custom-built analog projector for high-speed. With the high-speed prototype, we achieve accurate reconstructions at 450~fps and 1~Megapixel resolution. We recover the free-fall of a bunny at 450~fps with relative motion errors below 2\% (Figure~\ref{fig:teaser}).

Our contributions are: (1) A simple, effective, data-driven algorithm for structured light scanning that does not rely on optical projector calibration or triangulation; and (2) a hardware prototype comprising a digital camera and analog projector capable of acquiring accurate 3D geometry at 450~fps and 1~Megapixel resolution, or 1,450~fps at 0.4~Megapixel, demonstrating our
system's viability.

By rethinking the relationship of illumination, observation, and reconstruction, \systemname offers a new perspective on 3D scanning that is practical today and extensible for future developments.


\section{Related Work}
\label{sec:related}
While the broader literature on 3D geometry acquisition spans diverse solutions \cite{byo3d}, here we focus on SL approaches, which have been extensively surveyed, particularly for high-speed acquisition \cite{VanderJeught2016}.

\paragraph{Structured Light Scanning.}
Structured light (SL) scanning has a rich history, as reviewed by \cite{zhang_high-speed_2018}. Early systems, such as \cite{posdamer_surface_1982}, introduced the idea of projecting $m$-bit stripe sequences to encode $2^m$ patterns. This idea evolved through Gray codes \cite{hall-holt_stripe_2001, inokuchi1984range}, linear stripe patterns \cite{carrihill_experiments_1985}, RGB and color-encoded fringes \cite{boyer_color-encoded_1987,caspi_range_1998,je_colour-stripe_2012}, phase-shift techniques \cite{wust_surface_1991,gupta_micro_2012}, and space-filling curves \cite{gupta_geometric_sl_2018}.
Although effective for static scenes, these methods face limitations in dynamic, high-framerate settings. Most require projecting and decoding multiple time-varying patterns to establish correspondences for triangulation, which fundamentally restricts temporal resolution. Our system, in contrast, bypasses triangulation and captures depth from as few as one single projected pattern plus a white reference, enabling significantly faster acquisition via a beam-splitter-based optical setup. 
SL has also seen widespread deployment in commercial products, for example, the original Microsoft Kinect \cite{noauthor_kinect_2023}, based on PrimeSense technology \cite{noauthor_primesense_2023}. In our experiments, we compare with three commercial depth sensors capable of dynamic scanning: Microsoft Kinect, Intel RealSense, and Photoneo MotionCam. We note that these sensors provide visibly worse reconstructions of simple dynamic scenes, such as a fan spinning at 5~Hz.\looseness=-1
\paragraph{High-Speed SL Scanning.} 
Numerous efforts have been made to improve frame rates of SL techniques through hardware modifications. These include adaptations to DLP projectors \cite{zhang_superfast_2010, Koppal-2012-120215}, stereo-projector hybrid systems \cite{Jones2006, Jones2009, TABATA2016, Hu2021}, and event-based cameras \cite{Cossairt2015, Huang2021}. However, most of these approaches demonstrate limited dynamic reconstruction, achieving up to 120~fps with a single projector, and often not accurate enough to recover physical properties from the scene.

Their performance is further constrained by the limitations of DLP projectors, which support very high refresh rates only in binary mode. Although DLP hardware can operate at kilohertz speeds, these rates are typically not available for 8-bit or 24-bit color projections, as manufacturers optimize for perceptual frame rates rather than machine vision. In contrast, our proposed method relies only on the ability to repeatedly project a fixed pattern without explicitly modeling or calibrating the projector.
\paragraph{LUT-Based Calibration.}
Lookup tables (LUTs) have been used in SL systems primarily as a tool for calibration rather than reconstruction. \cite{LUO20146} constructs per-pixel LUTs by sweeping a linear stage and mapping phase values to depth for phase-shifted patterns. Similar pixel-wise LUTs are employed in telecentric SL systems \cite{Chen2022} to model nonlinear phase-depth mappings with high precision. In underwater imaging, LUT calibration methods account for refraction across glass-water interfaces \cite{Zoraja2023}. In contrast to these approaches, \systemname treats LUT not as an intermediate calibration aid, but as the primary mechanism for depth inference, enabling per-ray reconstruction without explicit projector calibration or projector-camera triangulation.
\paragraph{Summary.} SL scanning methods use calibration to fit an idealized projector model: while compact, the fitted model will unavoidably introduce errors. To the best of our knowledge, ours is the first SL approach sidestepping modeling the projector, replacing it with a purely data-driven LUT. 

\section{\systemname Scanning}
\label{sec:lookup}
\begin{figure*}[t]
    \centering
    \includegraphics[width=\linewidth]{fig/pipeline_cleaner.pdf}
    \caption{\textbf{Method Overview}. Our pipeline consists of two stages: \systemname Calibration and \systemname Reconstruction. During calibration, we sweep a planar target with a linear stage in fine depth increments. At each step, we use normalized pattern image against a white flash to produce per-pixel RGB observations. These are assembled into a per-pixel lookup table (LUT) that encodes the color-to-depth mapping. Once calibrated, we scan unknown geometry by capturing the same pattern and white flash, normalizing them, and querying the LUT to assign depths via nearest-neighbor search in Euclidean distance.
    }
    \Description[Pipeline of LookUp3D]{Pipeline of LookUp3D is divided into CALIBRATION and RECONSTRUCTION. During calibration, linear stage moves as a pattern is projected on a planar target. This is then saved into a lookup table with the color and depth for each pixel. For reconstruction, the same pattern is projected onto an object of unknown geometry. Then, we look, per-pixel, for the closest color match that was stored in the lookup table.}
    \label{fig:pipeline}
\end{figure*}


Our scanning method relies on a
per-pixel lookup table (LUT) \( C_i(d) \) from scene depth \( d \) to observed color values at camera pixel \( i \). Once this unique mapping is established during calibration, depth can be inferred at runtime by value lookup (Section~\ref{subsec:reconstruction}). We acquire each \( C_i \) using a linear stage and a flat calibration board. By sweeping the stage through a volume while projecting a fixed color pattern, we record the color response at each depth for every pixel, building a per-pixel LUT from depth to color (Section~\ref{subsec:calibration}). 
Fig.~\ref{fig:pipeline} summarizes the full calibration and reconstruction pipeline, and we provide additional technical details in Section~\ref{subsec:implementation}.

\looseness=-1

\subsection{Reconstruction}
\label{subsec:reconstruction}

\systemname reconstruction method relies on decoding depth from a per-pixel pre-recorded lookup table (LUT). We define $C_i(d): \mathbb{R}^+ \to \mathbb{R}^3$, the mapping from depth $d$ to color for a given pixel $i$. \rev{For reconstruction to work without confusion, the calibrated mapping should be unique.} The scanning procedure is as follows: we project a color pattern onto an object of unknown geometry and capture an image of intensity $I$. Reconstruction compares each pixel in the measured color image $I$ with a per-pixel LUT. In its unoptimized form, the operation is a simple, embarrassingly parallel lookup to find the depth whose color is closest to the observed pixel $I_i$, or
\begin{equation}
d_i^* = \argmin_{d \in \mathcal{D}} \left\|C_i(d) - I_i\right\|_2,
\end{equation}
where the search is done offline on a discrete set of depths $\mathcal{D}$.
The residual  \( r_i= \left\| C_i(d_i^*) - I_i \right\|_2 \) provides a per-pixel confidence measure that can be thresholded to filter unreliable points.

\looseness=-1

\subsection{Calibration}
\label{subsec:calibration}

The goal of calibration is to create the \rev{unique} dictionary $C_i\left(d\right)$ between depth and color for each pixel $i$. Our calibration procedure, then, uses a planar calibration target moved by an off-the-shelf linear stage \cite{Xiao_2015_CVPR} and a fixed pattern projected onto the calibration target. We record, per-pixel, the color reading and the depth measurement. First, in order to accurately retrieve a depth measurement from a camera pixel, we calibrate the intrinsic camera parameters following \cite{lanman_build_2009}. Then, we discretize the frustum in front of the camera as a collection of rays $\rB_i(d): \mathbb{R}^+ \rightarrow \mathbb{R}^3$, where $\rB$ departs from camera's origin and passes through pixel $i$, scaled by a positive depth $d$. At this point, we deviate from traditional structured light: to recover depth, we rely solely on the calibration board and the linear stage, without the need to model the projector \rev{defocusing, vignetting,} intrinsic and extrinsic parameters. We show in Section~\ref{subsec:static} a set-up where the quality of the optics of the projector does not alter the results achieved by \systemname, whereas a SL method reliant on triangulation suffers a lot in quality of reconstruction. 

\looseness=-1

\subsection{Technical Details}
\label{subsec:implementation}


\systemname is a data-driven method: both calibration and reconstruction rely exclusively on RGB measurements acquired from the camera sensor. While the core idea of lookup scanning is conceptually straightforward, achieving robustly accurate performance in static setting and adapting to dynamic setting require careful design. In this section, we detail the key implementation details and algorithmic strategies employed to improve the quality of the calibrated LUT and of the reconstructed 3D geometry.

\paragraph{Normalization.} Throughout, rather than work with raw readings, which are susceptible to varying albedo, lens vignetting, object texture, and other nuisance factors, we use normalized color \cite{caspi_range_1998}. To normalize our readings, we require one image acquired with the color pattern ($I_P$), one image of white flash ($I_W$), and one of ambient capture ($I_B$). The normalized color image is $I = \frac{I_{P} - I_{B}} {I_{W} - I_{B}}$. 
In dynamic, we forgo the ambient capture. This leads to a frame rate of half of projector and camera in the dynamic setting. We note that this is not equivalent to the superior option of normalizing to an ambient capture, which we use in static case.

\paragraph{Linear Stage Trajectory Fitting.}
During calibration, we detect the markers on the calibration board for every linear stage step, which allows us to register the camera in relation to the board plane. However, such registration is not entirely reliable due to marker detection errors and we benefit from incorporating the known, precise motion of the linear stage. We thus fit a linear trajectory to the calibration board motion using the stride size to acquire a monotonic depth measurement for each pixel.
We set up our calibration for $750$ linear stage steps of $0.2$\,mm, which in turn limits the depth precision per pixel at $0.2$\,mm. The calibration with these settings takes less than 10 minutes and requires no user interaction.
\paragraph{Denoising.} In low-noise settings, storing the normalized color readings into the LUT is sufficient, but when noise is significant (e.g., short exposures, low-quality camera sensors), denoising becomes a useful tool to improve depth decoding. In our high-dynamic case, where our method is more susceptible to sensor noise, we denoised the LUT via \rev{rolling kernel, Fourier transform, low-rank approximation, and spline fitting (refer to Section~\ref{subsec:denoising} for details).}
%
%
%
In our dynamic experiments, LUT storage is around 7GB uncompressed; with a low-rank approximation the storage reduces to 152MB.
\paragraph{Residual Thresholding.} The residual measures the consistency between the captured color and what was observed and stored in the LUT during calibration. Although a low residual does not guarantee correctness (e.g., due to LUT aliasing or local ambiguities), a high residual strongly indicates reconstruction failure (see Section~\ref{sec:conclusion}). Residual thresholding can be used to filter unreliable depth estimates, which is particularly beneficial for downstream tasks requiring high precision, such as physical property inference.

\paragraph{Restricting Depth Range.}  \rev{In our static set-up, a naive reconstruction with high channel count yields low-error results. With lower pattern count in both static and dynamic set-ups, however, we can opt for a coarse-to-fine (C2F) approach to effectively add neighboring information as we decode depth (Fig.~\ref{fig:c2f}). C2F helps disambiguate potential confusion from finding the minimum residual in a smaller depth range (refer to Section~\ref{subsec:c2f}). In dynamic scenes, we have additional access to the previous depth as a starting point for the lookup search; we see in practice that both C2F and temporal consistency (TC, refer to Section \ref{subsec:tc}) help disambiguate candidate depths in challenging dynamic cases (Fig.~\ref{fig:static_to_dynamic})}. \looseness=-1

\begin{figure}[ht!]
    \includegraphics[width=\columnwidth]{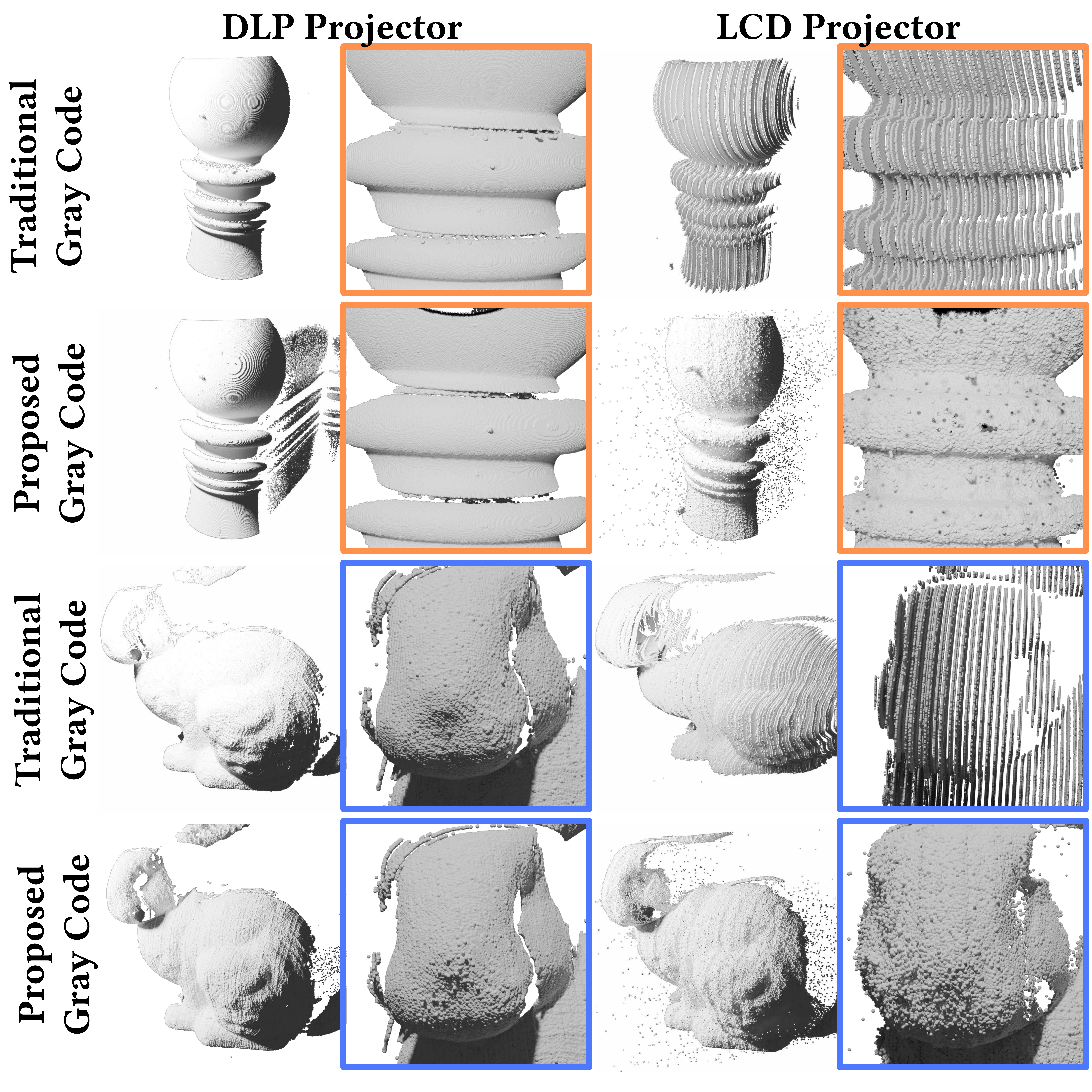}
    \caption{{\bf Static Reconstructions with Different Hardware.} We reconstruct a pawn and a bunny using a high-end DLP projector and an inexpensive LCD projector, both paired with the same camera. Traditional SL reconstruction with 44 Gray Codes deteriorates significantly with the low-quality projector due a reliance on accurate triangulation, whereas \systemname with 11 Gray Code channels remains robust.}
    \Description[Comparison Between Projectors]{Traditional Structured Light methods rely on triangulation between a calibrated camera-projector system. When the projector has poor calibration and is hard to model, Traditional Structured Light methods suffer -- the quality of the reconstructed pawn and bunny are much worse with a cheap LCD projector. LookUp3D, the method we propose, is largely unaltered by bad optics -- the pawn and bunny look similar with the good DLP projector and the bad LCD projector.}
    \label{fig:static_comparisons}
\end{figure}

\subsection{Static Validation}
\label{subsec:static}


We first validate \systemname in a static setting, where we minimize sources of noise and can more easily compare to traditional SL methods. We explore different hardware configurations, pattern combinations, \rev{and ambient light conditions}, conducting careful quantitative evaluations against ground-truth geometries that are otherwise difficult to obtain in a dynamic setting.

We conduct validation using two hardware setups, both sharing the same camera -- a Lucid Vision Labs Atlas 31.4MP with an Edmund Optics APS-C 50mm lens -- but paired with different projectors. The first is a high-quality Texas Instruments DLP4710EVM-LC projector. The second is a low-cost LCD projector (Mini Projector VOPLLS 1080P, approximately 50USD) intended for home entertainment with inexpensive optics. We control all other settings and compare the reconstructions across three test objects: the Stanford bunny (approx. 10cm $\times$ 12cm $\times$ 6cm), a precisely machined pawn (approx. 10cm $\times$ 16cm $\times$ 10cm), and a flat plane (40cm $\times$ 30cm). For the pawn, we use its CAD model as ground truth; for the plane, we assume ideal flatness; and for the bunny, we obtain a reference mesh using a traditional SL scanner and a rotating stage.

We evaluate \systemname using \rev{11,} 9, 6, and 3 channels and compare it to traditional structured light using Gray codes (with 44 binary patterns) with both projectors. We report the median absolute error in Table~\ref{table:static-comparison_alt2} and visually evaluate the reconstructions (Fig.~\ref{fig:static_comparisons}).

\begin{table}
\centering
\caption{Comparison of our proposed \systemname ({\it LU3D}) with traditional structured light. We report the root mean squared error (RMSE) in mm of each reconstructed point cloud with respect to the ground-truth geometry.}
\label{table:static-comparison_alt2}
\begin{tabular}{ p{0.8cm}p{1.0cm}p{0.7cm}p{0.6cm}p{0.6cm}p{0.7cm}p{0.6cm}p{0.6cm}}
 \toprule
 & & \multicolumn{3}{c}{DLP Projector} & \multicolumn{3}{c}{LCD Projector} \\
  \cmidrule(lr){3-5} \cmidrule(lr){6-8} 
 Method & \# Chan. &Bunny&Pawn&Plane&Bunny&Pawn&Plane\\
 \midrule
 Trad.\ & 44 
 &0.96&1.12&0.17
 &47.07&67.07&1.81\\
 {\it LU3D} & 3
 &1.46&1.86&0.42
 &3.02&1.42&0.42\\
 {\it LU3D} & 6
 &1.12&0.73&0.17
 &2.31&0.86&0.38\\
 {\it LU3D} & 9
 &1.03&0.34&0.13
 &2.01&0.63&0.34\\ 
 \bottomrule
\end{tabular}
\end{table}

\begin{figure*}[ht!]
    \includegraphics[width=.98\textwidth]{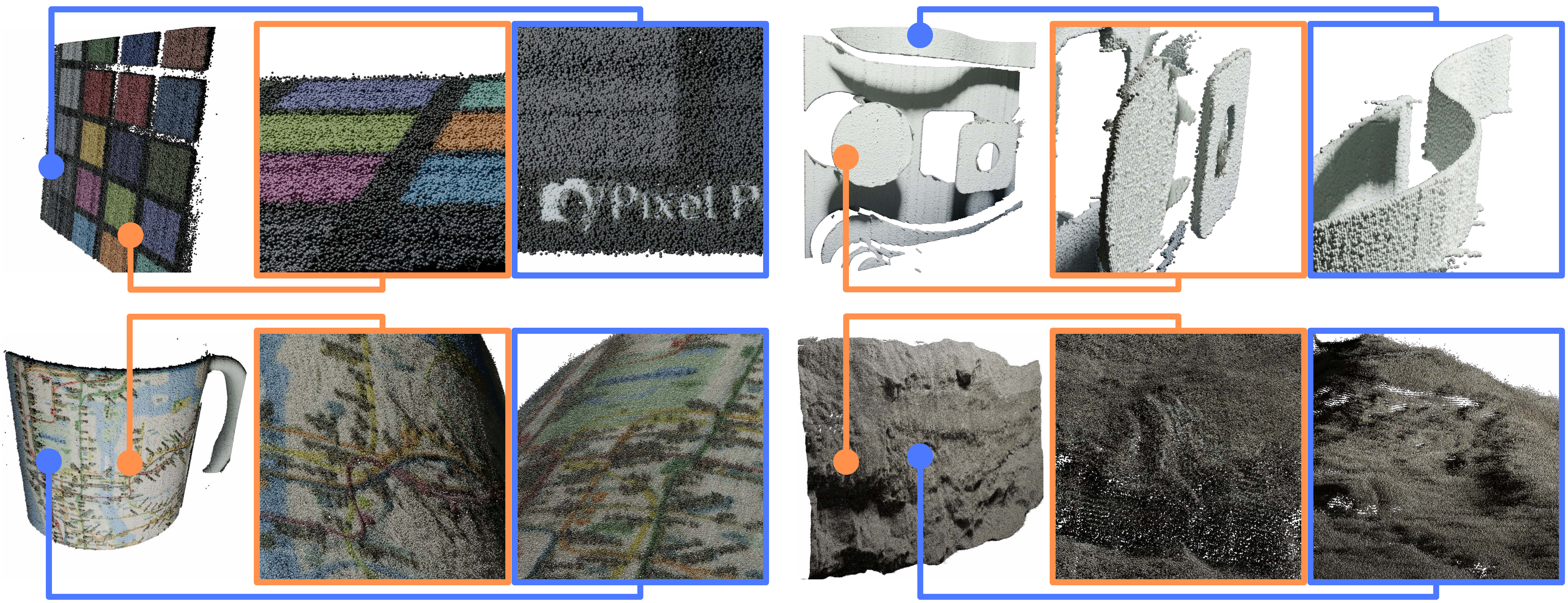}
    \caption{{\bf Mosaic of Static Scenes}. These scenes were captured with our DLP Projector and Atlas Camera in a static setting with controlled lighting to showcase the quality of reconstructions we can achieve with \systemname. We show four scenes reconstructed with 9 channels: top left is a color calibration board; top right is a precisely-machined irregular structure; bottom left is a mug; bottom right is a replica of a fish fossil. Each reconstructed cloud is displayed with two zoom-ins of different regions to highlight different shapes, textures, and albedo.}
    \Description[4 Reconstructions with LookUp3D]{This image shows details of four objects reconstructed with the proposed method LookUp3D. Even with varying shapes, textures, and albedo, the method is able to accurately reconstruct those details.}
\label{fig:static_figures_pages}
\end{figure*}

The results show that \systemname achieves reconstruction quality comparable to traditional SL with the DLP projector, and significantly outperforms it with the LCD projector. While the quality of traditional Gray Code reconstruction deteriorates significantly with the low-quality projector -- due to its reliance on accurate triangulation -- \systemname remains robust. We calibrated both projectors using the standard method in \cite{moreno_simple_2012}, but the calibration fails to sufficiently model the optical distortions in the inexpensive LCD projector. In contrast, \systemname implicitly embeds projector intrinsics into the LUT, so reconstruction accuracy is largely unaffected by projector quality. This insight motivates our design of the high-speed projector prototype (Section~\ref{sec:prototype}):  decoupling performance from the ability to model the projector optics enables us to build a simple analog projector with LED components.

Additionally, \systemname reconstruction error decreases with more pattern channels for both projectors. This validates the intuition that ambiguity in depth-to-color mapping decreases with increasing channel count. These results suggest that a 3-channel LUT is sufficiently accurate for our high-speed setting, where we trade off some accuracy for fewer patterns and higher fps. We defer an in-depth analysis of pattern design to prior work \cite{gupta_micro_2012, adaptive_color_structured_light, gupta_geometric_sl_2018, mirdehghan_optimal_2018} and our Supplemental Materials (Section~\ref{sec:pattern_design}).


We also reconstruct, with the DLP projector and \systemname, four challenging objects (Fig.~\ref{fig:static_figures_pages}): a flat color calibration board (top left), a precisely milled irregular structure, with curves and steps (top right), a mug (bottom left), and a fossil replica (bottom right). \rev{For the mug, we recovered a cylindrical diameter of $83.42$\,mm (versus $83.00$\,mm measured with a caliper) and RMSE of $0.68$\,mm; for the milled irregular structure, we got RMSE $1.35$\,mm from its CAD design.} For all four scenes, we also zoom into different parts of the reconstructed clouds to showcase that \systemname can capture those details like varying textures, varying colors, and difficult shapes.

\begin{figure}[hb!]
\centering
\includegraphics[width=.95\columnwidth]{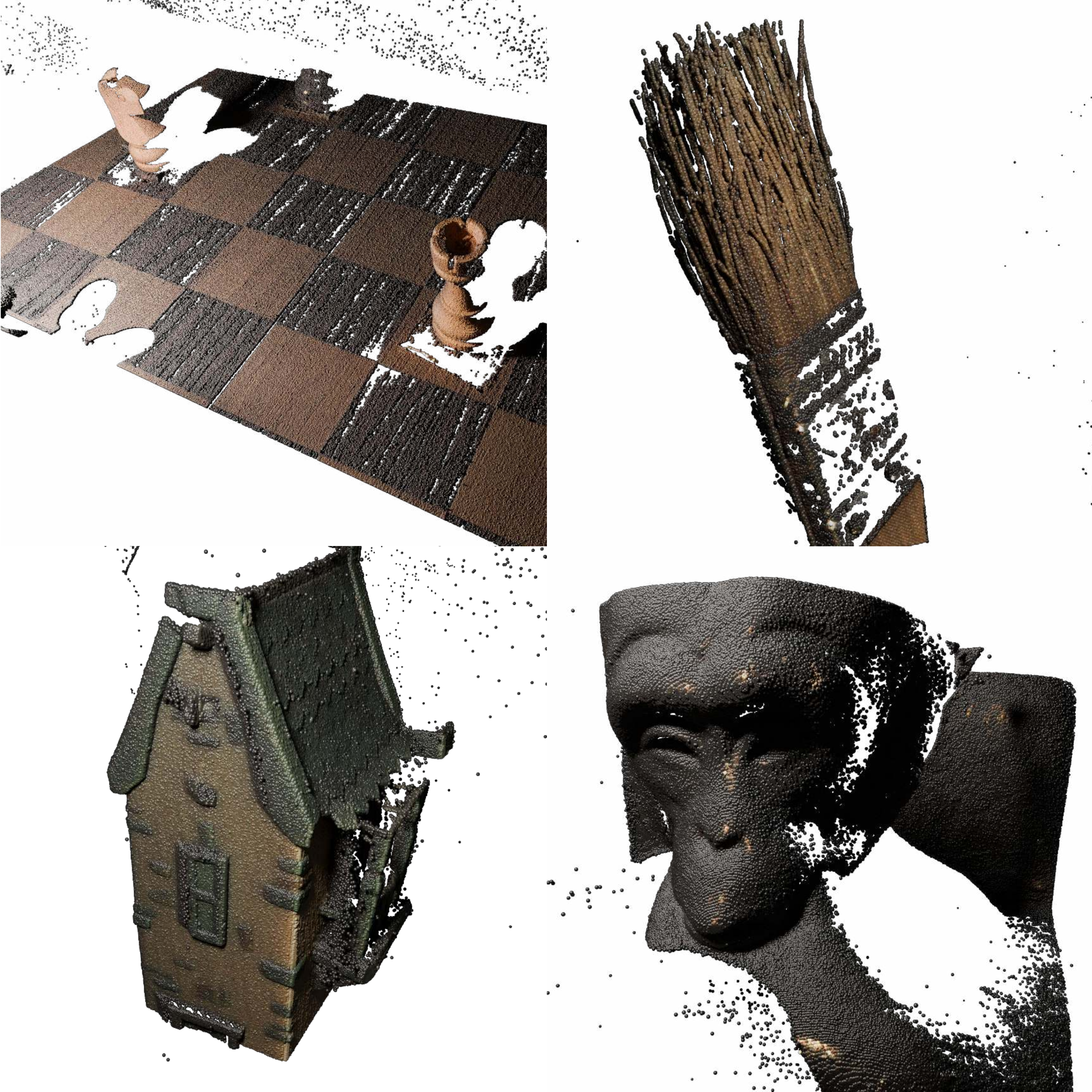}
\caption{{\bf Mosaic of Static Scenes with Ceiling Lights On.} These static scenes were captured with our DLP Projector with ceiling lights on. Even though the lookup table was calibrated in a darkroom, \systemname is able to accurately reconstruct the objects. We show a chess board, a paint brush, a 3D-printed house, and a monkey statue, all reconstructed with 11 channels.}
\Description[4 Reconstructions with LookUp3D with the ceiling lights on]{To show that the proposed method LookUp3D is robust, we reconstruct four objects, but now with the ceiling lights on. Even with the lights on, the method is able to accurately reconstruct most of the object.}
\label{fig:static_lights_on}
\end{figure}

\rev{Finally, we explore how \systemname performs under different lighting conditions. We calibrate an 11-channel LUT in a darkroom, but capture the objects with the ceiling lights on; the reconstructed objects maintain its accuracy even at a different lighting condition (Fig.~\ref{fig:static_lights_on}). The house is 3D printed, therefore we compare our reconstruction against the ground truth geometry and obtain median absolute error of $1.57$\,mm and RMSE of $2.46$\,mm. We provide the PLY files of twenty reconstructions with our Supplemental Materials.} 

The design of \systemname offers robustness and flexibility but introduces trade-offs. Building the lookup table requires a motorized linear stage for calibration, sufficient storage for the table itself, and enough RAM to support runtime lookup operations. However, these are one-time or amortized costs and do not impact the frame rate or simplicity of the scanning process during acquisition.

\section{High-Speed 3D Scanning}
\label{sec:dynamic}

\systemname opens the possibility of achieving high-resolution 3D scanning at high speeds. In this section, we review the hardware choices we made and present a prototype to scan 3D geometry at high speeds (Section~\ref{sec:prototype}). We also evaluate our method against controlled transformations of objects to which we have ground truth geometries acquired in a static setting with a traditional structured light scan (not \systemname) and extract physical properties, such as speed and gravitational acceleration, from the dynamic reconstructions (Section~\ref{sec:quantitative_analysis}). After establishing the accuracy of \systemname and our scanning prototype at high-speeds, we explore the new regime of 3D geometry scanning that \systemname enables. We highlight our capabilities with a handful of high-resolution, high-accuracy scenes reconstructed at 100~fps, 450~fps, and 1,450~fps (Section~\ref{sec:qualitative_results}). Finally, we compare \systemname side-by-side against three commercially-available dynamic 3D scanners: Photoneo MotionCAM 3D M+, Intel Realsense D455, and Microsoft Azure Kinect (Section~\ref{sec:commercial_comparison}).\looseness=-1

\begin{figure}[t]
    \centering
    \includegraphics[width=\columnwidth]{fig/static_to_dynamic_large_font.pdf}
    \caption{{\bf Adapting Reconstruction from Static to Dynamic Setting.} We illustrate key algorithmic improvements that enable reconstruction in the high-speed setting, demonstrated on a water balloon impact sequence. (a) The static reconstruction pipeline fails to recover many points with residual below $0.2$. (b) LUT denoising via low-rank approximation improves coverage but remains incomplete. (c) Coarse-to-fine (C2F) search helps recover locally consistent depths, further improving the completeness. (d) Adding temporal consistency (TC) leverages frame-to-frame coherence and helps reconstructing most of the object under challenging deformations.}
    \Description[Adapating Reconstruction from Static to Dynamic]{If we use the same naive reconstruction algorithm from the static set-up, our proposed method poorly reconstructs a deforming water balloon. However, if we add additional information, such as neighboring pixels and temporal consistency, LookUp3D can reconstruct most of the object even with challenging deformations at high-speed.}
    \label{fig:static_to_dynamic}
\end{figure}

\subsection{Prototype Hardware}
\label{sec:prototype}

\systemname's performance is bounded by the camera's resolution and frame rate, the projector's synchronized refresh rate, linear stage stride, and signal-to-noise-ratio permitted by the scene light. We built a prototype with as few constraints as possible to explore \systemname's capabilities with dynamic scenes.

We chose the high-end Chronos HD camera, which achieves 1k~fps at 2~Megapixel resolution and 3k~fps at 0.6~Megapixel, paired with a Sigma 35mm f1.4 lens for our high-speed scanner. 
%
%
%
Unfortunately, a similarly fast and sufficiently accurate projector is not readily available on the market. Commercially available DLP projectors, for instance, achieve high frame rates by trading off bit depth: the TI DLP LightCrafter 6500 projects 24-bit data at 500Hz, 8-bit at 1400Hz, 1-bit at 9000Hz.
To achieve high-speed capture, we require alternating between a RGB pattern (usually a 24-bit image) and a white flash. We propose a radically different hardware solution: an analog projector that can project the two alternating patterns at over 1k~fps and at high-resolution. 

\begin{figure}[ht!]
    \centering
    \includegraphics[width=0.65\columnwidth]{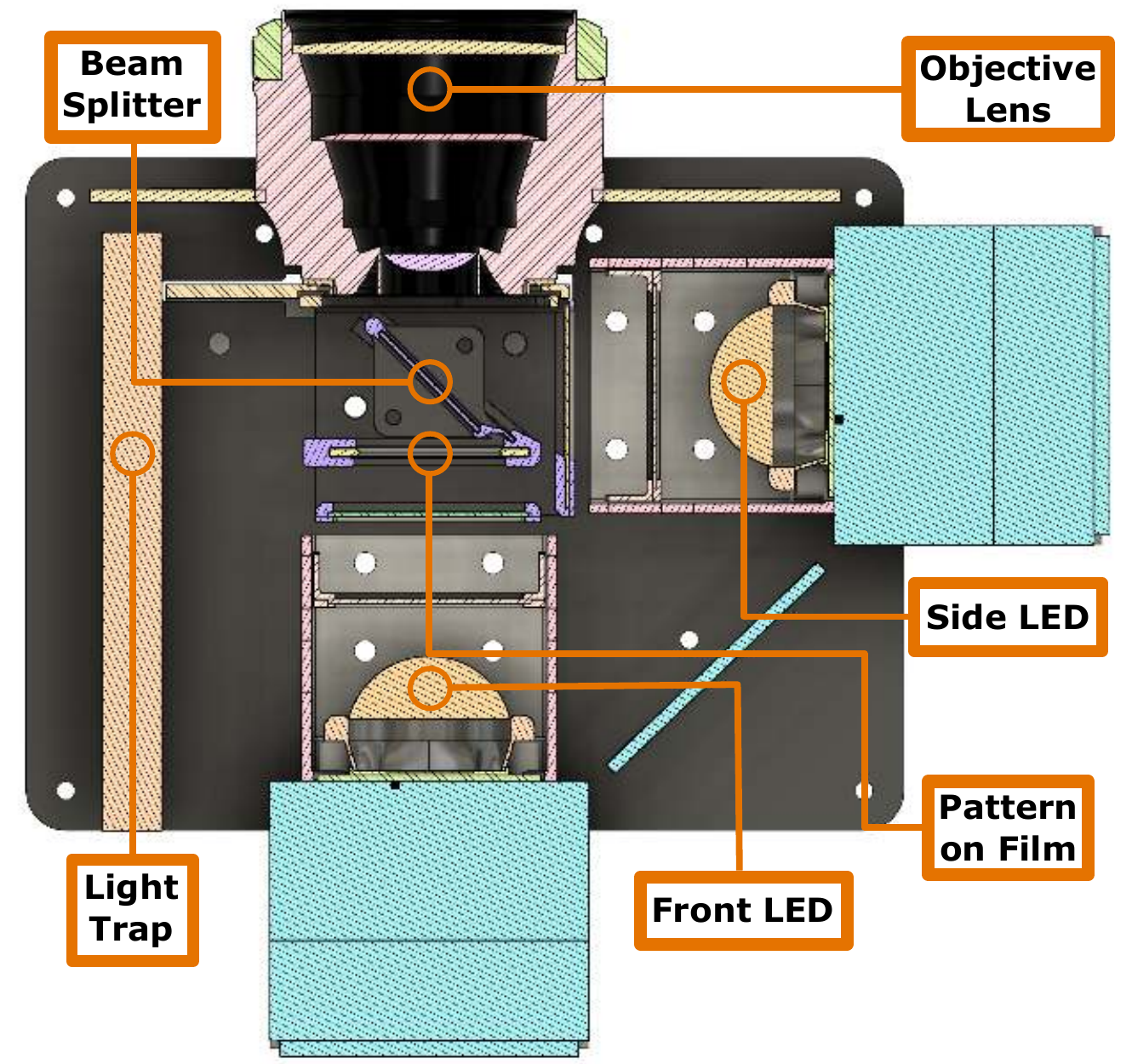}
    \Description[Analog Projector Diagram]{A computer-aided design model of the analog projector, which is proposed for the high-speed dynamic scenes in this paper. It highlights different pieces of the hardware, such as the pattern developed on film, the bright LEDs, a lens, and a beam splitter.}
    \caption{{\bf Analog Projector Diagram}}
    \label{fig:analog_projector}
\end{figure}

Our prototype analog projector (Fig.~\ref{fig:analog_projector}) comprises two 100~Watt LEDs perpendicular to each other. A beam splitter, placed at a 45-degree angle, ensures the flashes of light occur in the same frustrum in front of the projector. The projector is also equipped with a Sigma 50mm f1.4 lens. A 35-mm color film slide with the pattern exposed onto it is placed in front of one of the LEDs, which removes limits on bit resolution for the RGB pattern.

The projector house is made of laser-cut $3$\,mm-thick acrylic pieces. The holder for LEDs and the beam splitter were 3D printed with PLA. The lens holder was milled from a L-shaped piece of aluminum. Additionally, to ensure synchronicity between camera triggering and turning on and off LEDs at high frame rates, we use a Texas Instrument TIVA C Series TM4C123G ARM Cortex-M4F Launch Pad micro-controller; we can thus precisely switch between the color pattern and white flash at frequencies above 1~kHz. We are making the CAD design with all the individual pieces open source for the benefit of the community and to ensure reproducibility.

The only challenging part to reproduce in our setup is the film: we expose the film with a calibrated Full HD projector and a Canon EOS-1N analog camera with a Tamron 90mm f/2.8 lens. \rev{We document the procedure for bringing patterns onto film in the Supplemental Material (Section \ref{sec:film_development})}, and have stored 35mm film slides for shipping to labs interested in reproducing our setup.\looseness=-1

\subsection{Quantitative Analysis}
\label{sec:quantitative_analysis}

In this section, we test the accuracy and precision of \systemname by reconstructing and compare its 3D geometry against known geometries undergoing rigid transformations. We also test if \systemname is accurate enough to extract physical properties of dynamic scenes, since we think that a main potential application of \systemname is to to reconstruct and retrieve physical properties of objects undergoing rigid and/or soft transformations. This is applicable to robotics applications and to validating physical simulation models. 

\begin{figure}[t]
    \centering
    \includegraphics[width=\linewidth]{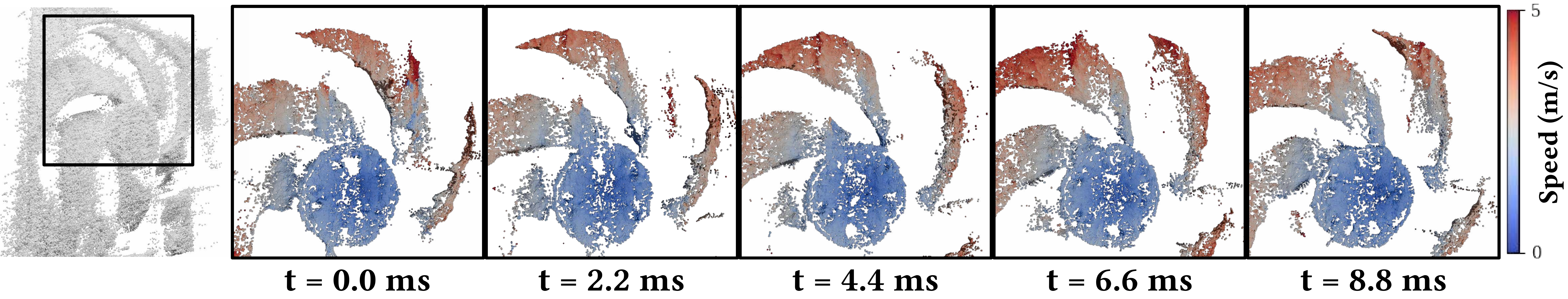}
    \caption{{\bf Fan Spinning Counterclockwise at 900~RPM.} We used our high-speed prototype with \systemname to reconstruct, at 450~fps, a fan. We focus on a zoomed-in part of the reconstructed fan and use a colormap to paint the speed (m/s) of each point in the cloud.}
    \Description[Fan Spinning at high speed]{A fan is displayed in several frames, with red color showing regions that move faster and blue color showing regions that most slower. As expected, we are able to extract higher speeds further away from the center of the fan, approaching 4 and 5 meters per second. In reality, these further points are predicted to be moving at 4.7 meters per second.}
    \label{fig:fan_1300rpm}
\end{figure}

\paragraph{Bunny Drop.} We drop a bunny onto a hard surface and reconstruct its impact at 450~fps (Fig.~\ref{fig:teaser}). More than capturing the shock deformation and oscillations of the bunny's ears, we can measure gravity from this sequence. We extract the centroid of 20 consecutive point clouds and fit a second degree polynomial to its displacement over time: we find a gravitational acceleration of $9.60$\,m/s\textsuperscript{2}, a $2\%$ relative error from the nominal value of $9.80$\,m/s\textsuperscript{2}.

\paragraph{Bunny on Linear Stage.} We then place the bunny on a platform and attach it to our linear stage. We set the linear stage to move 31.25\,mm and we reconstruct the bunny's motion at 450~fps (Fig.~\ref{fig:bunny_linear_motion}). With \systemname, we recovered a total distance traveled of 31.52\,mm, $0.9\%$ relative error from the ground truth, and a speed of $12.56$\,mm/s, whereas the linear stage manufacturer reports $12.50$\,mm/s.
\rev{To further evaluate the accuracy of reconstructing such geometry moving, we compare the error of our reconstructed bunny from the translated ground truth mesh (from Section~\ref{subsec:static}) and obtain a root mean squared error of $3.27$\,mm for $1,200$ frames.}

\rev{\paragraph{Fan Spinning.} We reconstruct a fan spinning at $900$\,RPM and find find point-to-point correspondences between adjacent frames using Iterative Closest Point. From these correspondences, we calculate the instantaneous speed for each point and display the reconstruction with a colormap in Fig.~\ref{fig:fan_1300rpm}. The furthest points on the blade are $0.5$ meters away from the center -- in circular motion at $900$\,RPM, the speed at that point would be $4.7$\,m/s. We extract speeds between $4.0$ to $5.0$\,m/s for the furthest points from the center.}

\rev{\paragraph{Ball Bouncing.} We grab a bouncy ball and perform drops on multiple materials to test collision elasticity. We reconstruct the drop and subsequent bounces of the ball to acquire coefficients of restitution for the different materials (Table \ref{table:ball_drop_experiment}). Since we measured the radius of the ball with a caliper ($31.00$\,mm), we also fit a sphere to the reconstructed model to verify the accuracy of our results. We track the sphere's center over time to find the velocity and gravitational acceleration (for which the nominal value is $9.80$\,m/s\textsuperscript{2}) of the scene. For the coefficient of restitution, we calculate $e= \left|\frac{v_{i}}{v_{f}}\right|$, where $v_i$ and $v_f$ are the the velocities {\it before} and {\it after} a collision, respectively.

\begin{table}
\centering
\caption{We drop a bouncy ball on multiple materials and reconstruct the scene with \systemname. We attempt to find the radius ({\it R}) of the ball and the gravitational acceleration ({\it g}) with our point clouds. We also calculate the coefficient of restitution {\it e} to evaluate how elastic the collision is.}
\label{table:ball_drop_experiment}
\begin{tabular}{ p{1.1cm}p{.9cm}p{1.3cm}p{0.8cm}p{1.4cm}p{0.8cm}}
 \toprule
 \multicolumn{6}{c}{Ball Drop Experiment}\\
  \cmidrule(lr){2-6} 
& Acrylic&Aluminium&Paper&Sand Block&Wood\\
 \midrule
 {\it R} (cm)  & 33.16 & 26.47 & 32.65 & 34.30 & 34.38\\
 {\it g} (m/s\textsuperscript{2}) & 10.99 & 9.57 & 11.03 & 11.90 & 10.28\\
 {\it e}  & 0.48 & 0.58 & 0.23 & 0.50 & 0.52\\
 \bottomrule
\end{tabular}
\end{table}
}

This opens the opportunity to integrate \systemname into applications in robotics and physical simulation models, where one can use the reconstructed point clouds to estimate physical properties.\looseness=-1

\subsection{Qualitative Results}
\label{sec:qualitative_results}

We have established, through our quantitative results, that \systemname is able to capture accurate motion at high frame rates. Here we highlight a mosaic of dynamic scenes (Fig.~\ref{fig:dynamic_scenes_100fps} at 100~fps and Fig.~\ref{fig:dynamic_scenes_450fps} at 450~fps) captured with our prototype and reconstructed with \systemname. Finally, we push our system to record a water balloon bouncing at 1,450~fps and 0.4~Megapixel (Fig.~\ref{fig:dynamic_scenes_1450fps}) -- even in a controlled lighting environment, sensor noise is really noticeable with camera exposure at 0.3\,ms and \systemname's performance decreases. We also stress that, due to the high speed in capture and the deformations of the materials in the scenes, there is no ground truth available and we do not report any error metric for these reconstructions. \looseness=-1

\subsection{Commercial Comparisons}
\label{sec:commercial_comparison}

\begin{figure}
    \centering
    \includegraphics[width=\columnwidth]{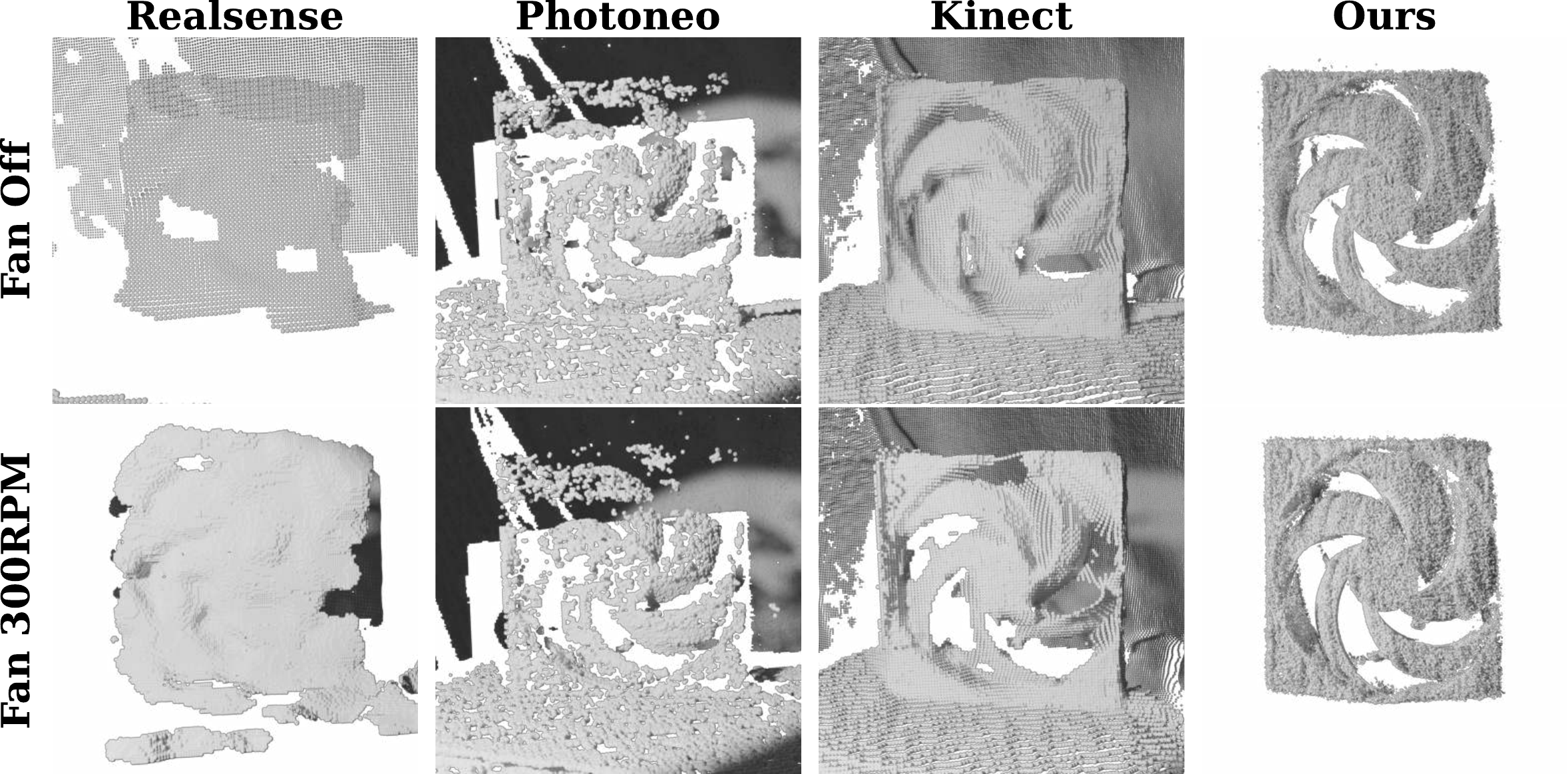}
\caption{{\bf Comparison of Reconstructing a Fan.} Top: an inactive fan captured by three commercial devices (Relsense, Photoneo, Kinect) and our high-speed scanning prototype. Bottom: the fan spinning at 300~RPM. The video in the Supplemental Materials shows a video of each system.}
\Description[Comparing a Fan with Different Commericial Devices]{The Realsense, Photoneo, and Kinect cannot capture the geometry nor the speed of the fan spinning at 300 rotations per minute. Our method, on the other hand, is able to get accurate, high-resolution data as the fan spins fast.}
    \label{fig:fan-comparison}
\end{figure}

In addition to the quantitative evaluations, we compare our high-speed prototype against three commercial dynamic depth sensing devices: Photoneo MotionCAM 3D M+, Intel Realsense D455, and Microsoft Azure Kinect. These scanners cannot be used simultaneously as they all actively emit light; we thus repeat each scene for each scanner, striving to keep the initial conditions as similar as possible. We note that there is a discrepancy in the frame rate and resolution of the scanners: 8~fps/0.9MP Photoneo; 90~fps/0.04MP RealSense; 30~fps/0.4MP Azure Kinect; and 450~fps/1MP for ours.
For better comparison, we first use the devices to capture a fan turned off (top row of Fig.~\ref{fig:fan-comparison}). Then, we turn on the fan to 300 RPM and visually evaluate each scanner's performance (bottom row of Fig.~\ref{fig:fan-comparison}). The {\it RealSense} device we tested makes use of stereo-matching -- it is equipped with two infrared cameras and it projects an infrared pattern to assist in feature matching across the two different views. It has the highest frame rate (90~fps), but a low resolution point cloud. {\it Photoneo} makes use of parallel structured light with a proprietary CMOS sensor. It offers the highest resolution, but at a very low frame rate (8~fps). The {\it Kinect}, equipped with a Time-of-Flight camera, strikes a balance between the previous two scanners in terms of scanning speed (30~fps) and resolution. Our scanner, in contrast, compromises on neither scanning speed nor resolution. 

\begin{figure*}
\centering
\includegraphics[width=.95\textwidth, trim=20pt 0pt 0pt 0pt, clip]{fig/dynamic/figures_page_100fps.pdf}
\caption{{\bf Dynamic sequences for different scenes at 100~fps.} The exposure for each captured frame was 4.5 milliseconds. For the white capture images, we synthetically increase gain and contrast, otherwise the captures are too dark to be properly displayed. We also add boxes on the reconstructed point clouds to aid visualization of subtle motion. The first row is a hand moving; the second row is a piece of cloth being twisted; the third row is a silicone cube being squeezed; the fourth and last row is a 3D-printed piece being bent from its edges. We have provided clips of the scenes in our accompanying video.}
\Description[Reconstructions at 100 frames per second]{LookUp3D is used to reconstruct a moving hand, a twisting piece of cloth, a cube being squeezed, and piece of PLA being bent. For each scene, there are 5 point clouds displayed, being 90 milliseconds apart from one another. This shows our proposed method, combined with our analog projector, can capture dynamic scenes with high quality at 100 frames per second.}
\label{fig:dynamic_scenes_100fps}
\end{figure*}
\section{Concluding Remarks}
\label{sec:conclusion}

We introduced a new approach for structured light 3D scanning that does not require modeling or calibrating the projector, and demonstrated its applicability for both static and dynamic scenes.

\begin{figure}[hb!]
    \centering
    \includegraphics[width=\columnwidth]{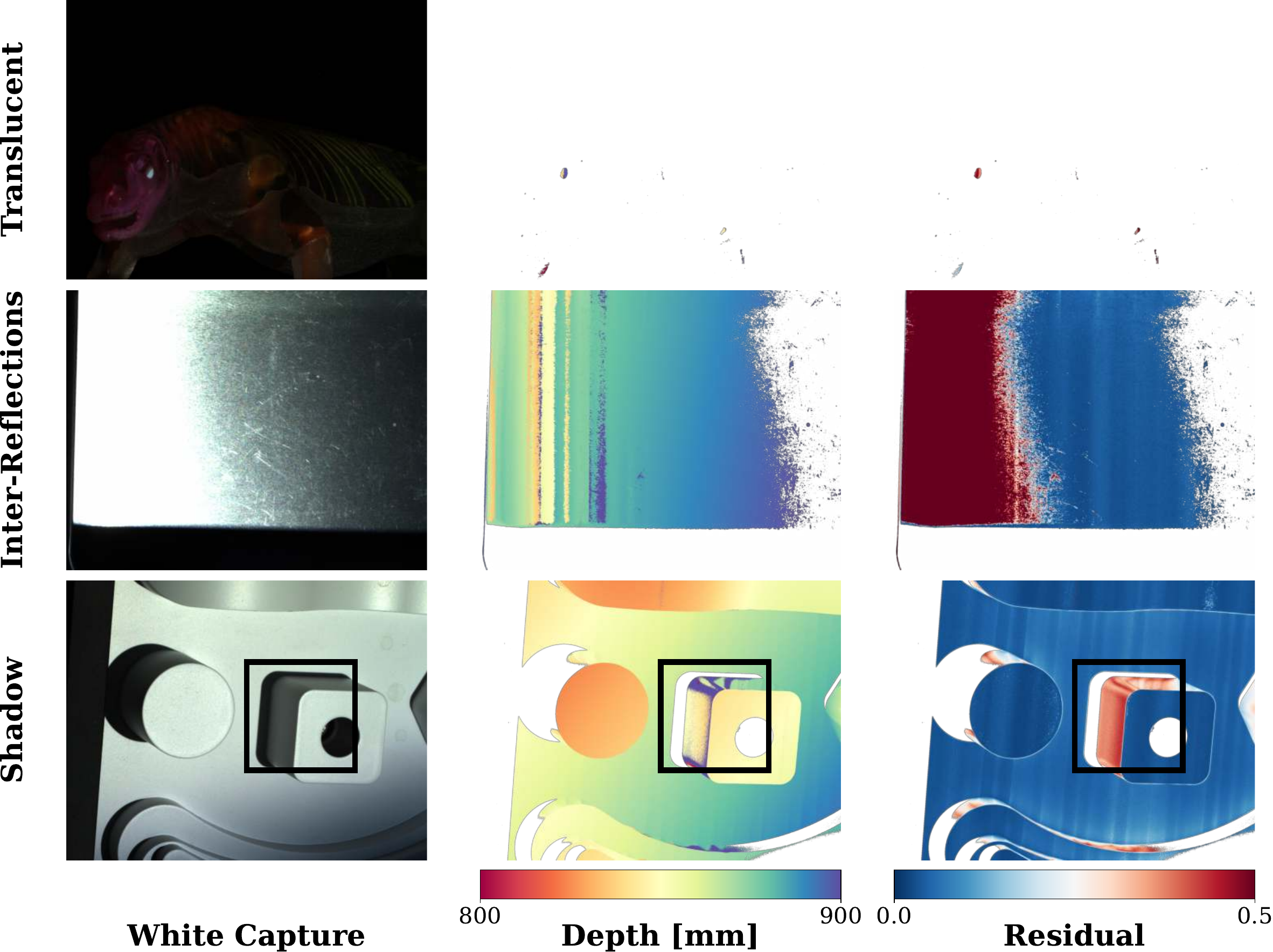}
\caption{{\bf Limitations.} \systemname is designed to record light that diffusely scatters from a calibration board. When light refracts through a translucent medium, our method has no chance of working. Similarly, we cannot decode overexposed pixels or inter-reflections. In the bottom row, the square highlights a shadowed region where decoding fails.}
\Description[Failures of Proposed Method]{Our proposed method, LookUp3D, fails for reconstructing a transparent lizard toy, a metallic lunch box with several inter-reflections, and the detail under heavy shadow of an object.}
    \label{fig:failures}
\end{figure}

\paragraph{Limitations.} Since we rely on light diffusely scattering off of the calibration board, our system cannot resolve inter-reflections, translucent objects, highly specular or reflective surfaces, and shadow regions (Fig.~\ref{fig:failures}), similarly to other structured light methods. \rev{Additionally, \systemname reconstructions are inherently restricted to the calibrated depth range, as there is no direct way to extrapolate the data to regions outside of the volume. We also attempted to use our high-speed prototype with \systemname to reconstruct a balloon popping (Fig.~\ref{fig:balloon_popping}). Unfortunately, even at 1450~fps, the motion of the explosion is too abrupt: since we acquire two images (a pattern projection and a white flash), if these two images do not overlap much, \systemname will most likely fail to recover any relevant depth. Finally, our projectors have limited light power and our method will not work if the flux of ambient light is much higher than that of the projector (e.g., outdoors). For more details on ambient light in our dynamic set-up, refer to our Supplemental (Section \ref{sec:ambient}).}

We also notice that wrong depths align well with high residuals, showing their utility as an indicator of reconstruction confidence. The ease of calibration of the system might allow for some creative solutions for these challenging cases, such as adding multiplexed light sources and calibrating with a specular pattern at different orientations, which we believe are exciting avenues for future work.

Another limitation is that our reconstructions are currently done offline, whereas the commercial devices provide real-time data. Our current unoptimized implementation takes 2 seconds per frame on a CPU (1 core of an AMD EPYC 9654P Processor clocked at 3.7 GHz) \rev{and 0.1 second per frame on an NVIDIA RTX A6000 GPU}.



\paragraph{Future Work.} 
Increasing the LED power (and thus the amount of light available) from our analog projector would allow us to achieve even higher frame rates: we currently cannot further increase the power output as our cooling system is unable to dissipate it, leading to a melted film. We believe that the simplicity offered by our approach (a fixed image projector) will foster additional hardware research in this direction, overcoming this limitation, and is a much easier proposition than developing high frame rate projectors for standard SL pattern stacks. A straightforward, but challenging from an engineering standpoint, way to increase power would be using a larger film, but other options such as printing on or etching a glass slide might lead to a more compact projector. If this limitation is overcome, we expect to acquire depth data of similar quality as our static scans but for dynamic scenes. We note that there is only one difference between our static and dynamic results, the exposure time of the projector and camera, which directly affects sensor noise and reconstruction errors.

The possibility of using arbitrary patterns with our \systemname reconstruction algorithm and the removal of the necessity to calibrate the projector opens many interesting avenues for future work: (1) it is possible to use multiple synchronized projectors and cameras to increase scanning view without worrying about interference (as long as the projectors do not overexpose), (2) the use of multiple analog projectors multiplexed in time to reduce noise at the cost of lower temporal resolution, (3) the use of invisible light patterns (infrared) potentially combined with visible light.

\begin{figure}[t]
    \centering
    \includegraphics[width=\columnwidth]{fig/residual/residual_in_future_work.pdf}
    \caption{{\bf Residual Filtering and Depth Requery.} (a) The original pawn result has artifacts around its neck, caused by heavy shadows, specularities, and inter-reflections. (b) Filtering at a threshold of 0.05 removes many wrong depths. (c) A stricter threshold of 0.02 removes noticeable outliers but yields a less complete point cloud. (d) Using the inliers from (c) to restrict the search range leads to a more complete and accurate reconstruction.}
    \Description[Removing Poorly Reconstructed Regions]{With the use of residuals, we can remove points that don't meet a certain threshold. In the figure, we show, with red, points that have a high residual in the pawn object. Most of these points are clearly erroneous. As we decrease the threshold, we remove more and more points from the point cloud; although the remaining point cloud is sparser, its accuracy is much higher.}
    \label{fig:residual_future_work}
\end{figure}

Another promising direction lies on the algorithmic side, where we can take advantage of two key properties of the existing system: the availability of per-pixel residuals as a natural byproduct of the reconstruction process, and the structured, differentiable nature of the underlying depth-to-color mapping. Residual maps not only serve as confidence indicators but also enable refinement on depths. In a preliminary experiment on the static pawn scene using \systemname with only three channels, shown in Fig.~\ref{fig:residual_future_work}, we identify outliers via residual thresholding and requery their depths using a narrowed search range guided by depths from nearby inliers. Without additional input or parameter tuning, we significantly improve the reconstruction accuracy in the same challenging regions highlighted in our failure cases -- such as specularities, inter-reflections, and heavy shadows, reducing the root mean squared error (RMSE) from 6.39mm to 1.52mm. We believe residual-based refinement holds strong potential for further improving robustness and accuracy, especially in dynamic or noisy settings. Building on this, a second direction involves embedding the lookup into an optimization framework. In particular, using residuals to guide (stochastic) gradient descent for depth refinement
 could lead to more accurate and stable reconstructions.
 
\paragraph{Reproducibility.} 
To foster reproducibility and adoption of this new technique, which we believe offers a different and exciting perspective on structured light scanning, we will release an open source implementation\footnote{ \url{https://github.com/geometryprocessing/scanner}} of our algorithm and an open hardware specification for our high speed projector.\looseness=-1
\section*{Acknowledgments}
This work was partially supported by the the NSF grants OAC-2411349 and OAC-2411221. 
Giancarlo Pereira was partially supported by the New York University Tandon School of Engineering Fellowship.
We thank NYU IT High Performance Computing services, for help with resources, services, and expertise.
We thank Christopher Musco for fruitful discussions regarding low-rank approximations.


\clearpage
\clearpage



\begin{figure*}
\centering
\includegraphics[ width=.95\textwidth]{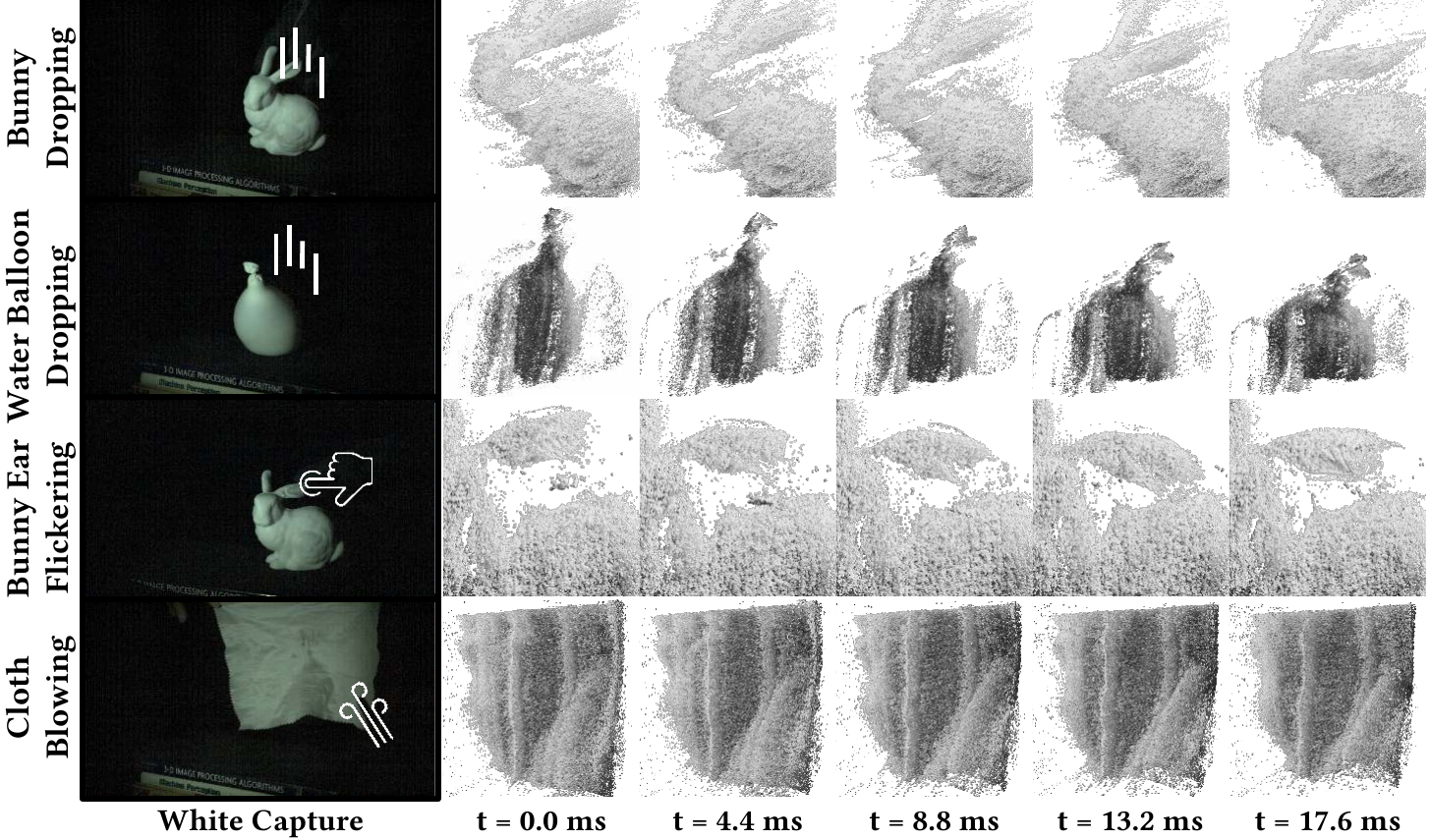}
\caption{{\bf Dynamic sequences for different scenes at 450~fps.} The sensor exposure was set to 1 millisecond, which decreases signal-to-noise ratio and causes more sporadic noise in the reconstructed point clouds. For the white capture images displayed above, we synthetically increase gain and contrast, otherwise the captures are too dark to be properly displayed. The first row is a silicone bunny in free fall hitting a hard surface; the second row is a water balloon in free fall also hitting a hard surface; third row is the silicone bunny's ear wiggling at high frequency; the fourth and last row is a cloth hanging while air blows on it. We have provided clips of the full scenes in our accompanying video.}
\label{fig:dynamic_scenes_450fps}
\Description[Reconstructions at 450 frames per second]{LookUp3D is used to reconstruct a bunny in free fall, a water balloon bouncing on a hard surface, the bunny ear's being flicked, and a piece of cloth being blasted with wind. For each scene, there are 5 point clouds displayed, being 4.4 milliseconds apart from one another. This shows our proposed method, combined with our analog projector, can capture minute movement in dynamic scenes with high quality at 450 frames per second.}
\end{figure*}

\begin{figure*}
\centering
\includegraphics[width=.95\textwidth]{fig/dynamic/figures_page_1450fps.pdf}
\caption{{\bf Sequence at 1450~fps.} We highlight the motion of a water balloon as it hits a hard surface in less than one twentieth of a second. To achieve 1450~fps, we decrease the camera resolution to 0.4Megapixel and expose each frame for 0.3 millisecond. Sensor noise is very pronounced at this exposure and decreases the points we can confidently reconstruct. The bottom row are zoom-ins of consecutive frames as shock wave propagates to the top of the water balloon. This exemplifies that \systemname captures minute details in very small time steps. We have provided a clip of the scene in our accompanying video.}
\Description[Reconstructions at 1450 frames per second]{LookUp3D is used to reconstruct a water balloon bouncing on a hard surface. We capture the shock wave propagating on the surface of the balloon, with frames reconstructed just 0.67 milliseconds apart from one another. With such low exposure, noise is heavily increased in these point clouds, but our method can still capture some of the points undergoing deformation in less than one twentieh of a second.}
\label{fig:dynamic_scenes_1450fps}
\end{figure*}

\clearpage
\bibliographystyle{ACM-Reference-Format}
\bibliography{99-bib}

\clearpage
\setcounter{page}{1}
\appendix
\section{Supplemental Material}
\label{sec:supplemental}

The sections below form the Supplement text to the paper "\systemname: Data-Driven 3D Scanning." We cover the effects of ambient light (Section \ref{sec:ambient}), the effects of over a large volume of reconstruction (Section \ref{sec:defocusing}), additional details on residual analysis and thresholding (Section \ref{sec:residual}), additional details on pattern design and evaluation (Section \ref{sec:pattern_design}), and the film development for patterns (Section \ref{sec:film_development}). We also explore more implementation details of \systemname, including denoising of the lookup tables (LUT), the use of a coarse-to-fine (C2F) approach in reconstruction to effectively add neighboring information, temporal consistency (TC) in dynamic scenes to restrict the depth range, and the exciting avenue of using a multilayer perceptron to learn the depth-to-color mapping (Section \ref{sec:lookup_details}). We also display a few additional dynamic reconstructions in Section \ref{sec:additional_dynamic}.

\section{Effects of Ambient Lighting}
\label{sec:ambient}

\begin{figure}
\centering
\includegraphics[width=\columnwidth]{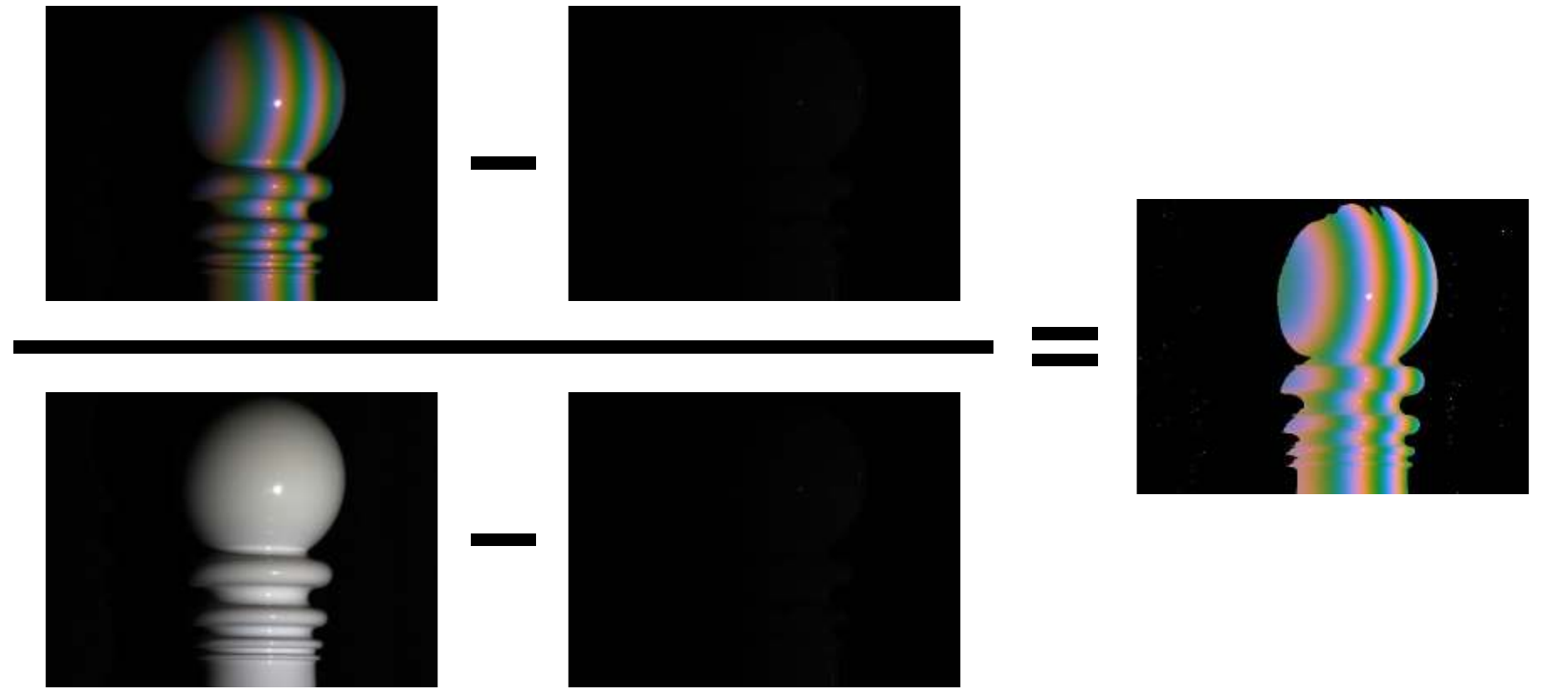}
\caption{Normalized colors (right) are obtained by dividing each pixel of an image acquired while projecting a pattern (top left) with the corresponding pixel of an image obtained by projecting white (bottom left), subtracting an image obtained while projecting black (middle) from both. The subtraction of ambient image becomes negligible in a darkroom, which we exploit in our high-speed setting to increase frame rate.
\vspace{1pt}}
\Description[Normalization]{We show what the pawn looks like in three conditions: with a color pattern projected on it, with a white flash, and with nothing being projected (we call it ambient). To get the normalized color of the projected pattern, we subtract ambient from both white and pattern, and then divide, pixel-wise, pattern by white to obtain an image with albedo and textures removed.}
\label{fig:normalization}
\end{figure}

Normalization is an important step in image processing to remove varying albedo, texture, and other nuisances in dealing with capturing objects. We use three different captures in \systemname to construct the normalized pixel intensity of a capture: a pattern illuminated $I_P$, white-flash illuminated $I_W$, and ambient non-illuminated $I_B$. The procedure, then, is element-wise described by the equation

\begin{equation}
    I = \frac{I_P - I_B}{I_W - I_B}
\end{equation}
and displayed in Fig.~\ref{fig:normalization} with the example of a pawn.

In the high-speed setting, however, reducing the number of captures is important for achieving high frame rates, avoid motion blur, and accurately capturing the motion or deformation of objects.

\rev{In the main dynamic results presented,} we forgo the ambient subtraction, which enables frame rates that are $1.5$ higher in capturing dynamic scenes. We chose to capture all our high-speed data with the room blinds down and with the ceiling lights off, to get the best quality data out of our method and prototype. Furthermore, making the analog high-speed projector significantly brighter (using 100~W LEDs) is not only reducing the image noise at shorter exposures but also contributes to a $I,I_{W} >> I_{B}$ relation that further reduced the impact of ambient light in a darkroom.

This does bring in a limitation: the calibration and the data capture should be taken with similar lighting conditions, otherwise the quality of reconstruction degrades. For instance, if the calibration is done with the ceiling lights turned off, the reconstruction captures should not be done with ceiling lights on to avoid adding confusion into the depth decoding step. 

\rev{While these main dynamic results were taken in darkroom for simplicity and reproducibility, we confirm that LookUp3D can obtain dynamic results with ambient light: we capture and subtract an ambient image at each timestep. This comes at the cost of a decreased frame rate, for instance from 450fps to 300fps. We highlight in this section two scenes: a hand wiggling its finger and a ball bouncing. Both scenes had the LUT calibrated in a darkroom, but the scenes were captured with the ceiling lights on.

\begin{figure*}[t]
    \centering
    \includegraphics[width=0.99\linewidth]{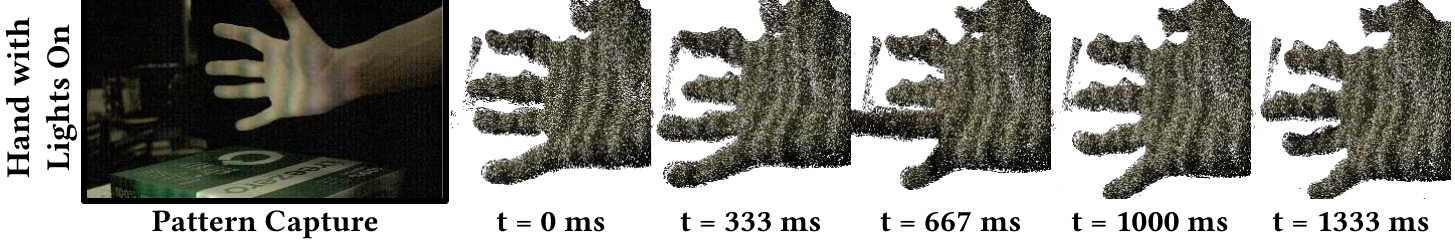}
    \caption{{\bf \systemname Reconstruction with Ceiling Lights On.} We reconstruct, at 300~fps and $1$\,ms exposure, a person's hand wiggling their fingers. To handle ambient light, we now take three images per reconstructed frame: one with LED pattern on, one with white LED on, and one with both LEDs off.}
    \label{fig:hand_lights_on}
    \Description[Dynamic Reconstruction with Lights On]{We reconstruct a dynamic scene with the ceiling lights on: hand wiggling its fingers. There are 5 point clouds displayed, being 333 milliseconds apart from one another. This shows our proposed method, even with lights on, can capture in dynamic scenes, as long as we also capture an ambient image.}
\end{figure*}

Similar to the experiment in Section~\ref{sec:quantitative_analysis}, we drop a ball on top of a ream of paper and let it bounce. For each reconstructed frame, we fit a sphere to the point cloud -- this returns both the radius of the sphere and the residual (summed Euclidean distance of all points from perfect sphere). We use both values to calculate a weighted mean (using the inverse of the residual as the weight) of 31.20mm for the ball radius, whereas the real life measurement with a caliper is 31.00mm.

We additionally measured the maximum flux (in a darkroom) of our analog projector with a light meter: placed 0.5m away, it measured 1,510 Lux; at 1.0m, 650 Lux; and at 1.5m, 210 Lux. We then switched the ceiling lights on (700 Lux) and attempted to calibrate and reconstruct dynamic scenes: at 1.5 meter away, our projected pattern is no longer discernible in the scene when the ceiling lights are on.}

\rev{\section{Effects of Defocusing}
\label{sec:defocusing}
\begin{figure*}[t]
    \centering
    \includegraphics[width=\linewidth]{fig/dynamic/bunny_defocus.pdf}
    \caption{{\bf Effect of Defocus as Bunny Moves Through Large Volume.} We present a larger depth range with \systemname to understand the impact of defocusing. On the left, we plot what a calibrated ray from the LUT looks like, ranging from 0.8m to 1.5m of depth -- any vignetting and defocusing effect from the projector gets baked into the LUT. On the right, we display three frames of bunny while the projector shines the {\it Spiral} pattern. Although we recorded this scene at 200~fps, we show three frames, each separated by $2.00$\,seconds, to highlight the bunny's traveled distance. Immediately below each image, we render the reconstructed point cloud. The bunny moved, in this time span, approximately $30$\,centimeters forward and \systemname is still able to reconstruct the bunny's geometry and position accurately.}
    \label{fig:bunny_defocus}
    \Description[Reconstruction in Larger Volume]{LookUp3D is used to reconstruct a a bunny in a larger volume. There are 3 point clouds displayed, covering about 30 centimeters in space. This shows our proposed method can capture scenes in a large volume of reconstruction, even under the effect of projector defocusing.}
\end{figure*}

With \systemname, we store the normalized RGB values directly into a lookup table; therefore, any effect due to defocusing or vignetting from the projector gets baked into the LUT. This makes the approach particularly appealing since we skip modeling any complex optical behavior, including projector defocusing.

For this experiment, we use our dynamic set-up and calibrate a LUT with $1000$ intervals, each with a linear stage stride of $0.5$\,mm -- this calibration still takes under 10 minutes and requires no user interaction. We plot, on the left-hand side of Fig.~\ref{fig:bunny_defocus}, a ray that covers around $70$ centimeters of depth, raging from $0.8$ meters to $1.5$ meters of distance from the camera sensor.

For the test scene, we grab the Bunny and move it through this large volume of reconstruction. The reconstructed point clouds in Fig.~\ref{fig:bunny_defocus} show that we can still reconstruct the bunny as it sweeps the volume, with more noise in the first frame due to a decreased signal-to-noise-ratio as the bunny is around $1.25$\,m away from the camera-projector pair. The distance of the bunny between the first and last displayed frames is approximately $30$ centimeters.}

\section{Residuals Analysis}
\label{sec:residual}
The residual, computed as $r_i = \left\| C_i(d_i^*) - I_i \right\|_2$ indicates consistency with the calibration data and serves as a proxy for reconstruction confidence. It is an inherent byproduct of our lookup-based reconstruction process. No additional models or heuristics are required to assess confidence, making the system naturally self-explanatory.

\begin{figure}
\centering
\includegraphics[width=0.9\columnwidth]{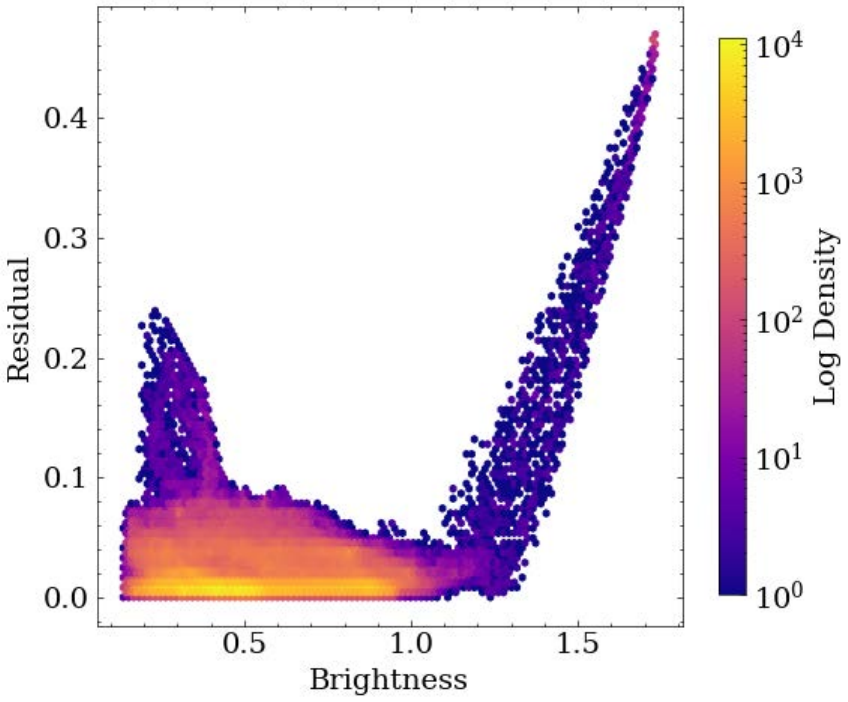}
\caption{Relationship between residual and observed brightness. This hexbin plot visualizes per-pixel density over residuals (y-axis) and pre-normalization brightness (x-axis). The residual exhibits a clear U-shaped trend: for brightness below 0.5, residuals increase moderately; above 1.0, residuals rise sharply in an almost linear fashion; and in between, residuals remain consistently low (under 0.1). Most points are concentrated in this stable region, suggesting that the majority of the scene is reconstructed with high confidence.
}
\Description[Residual]{For points in low-light and for points in almost overexposed regions, our proposed method suffers the most.}
\label{fig:residual_brightness}
\end{figure}

To further understand and validate this signal, we investigate the relationship between the residual and observed brightness of each pixel (represented as the l2
norm of its RGB vector before normalization). As shown in Fig.~\ref{fig:residual_brightness}, we observe a clear U-shaped trend: residuals are low for moderate brightness levels and rise at both extremes, which aligns with what we observe in the failure cases -- dark pixels (e.g., in shadows) suffer from low signal-to-noise ratio, while overexposed regions (e.g., specularities, inter-reflections), saturate and flatten RGB variation, weakening discriminability. This observation highlights that residuals reflect physical image formation properties, and that our system performs reliably with moderate brightness.

To evaluate the effectiveness of using the residual to identify incorrectly reconstructed points (i.e., residual-based filtering), we conduct experiment on the same static scene (the pawn) captured with the spiral pattern. We test multiple threshold levels and report quantitative metrics against the ground truth CAD model (Table~\ref{tab:residual_thresholding_eval}). While the main paper uses a coarse threshold of 0.1, progressively stricter thresholds yield sparser but more accurate reconstructions. This observation enables a trade-off: downstream applications requiring precise surface geometry (e.g., measurement or physical property inference) may benefit from aggressively filtered point clouds, at the cost of completeness. To address this, we present in the future work preliminary results on residual-based refinement, a promising direction for further exploration.



\begin{table}[ht]
\caption{Quantitative evaluation of residual-based filtering. Tighter thresholds lead to consistent drops in both median absolute error (MedAE) and root mean squared error (RMSE). MedAE remains more stable, as it is less sensitive to outliers. Notably, filtering at a threshold of 0.02 yields an RMSE of 1.53~mm compared to 6.39~mm without filtering, by removing 23\% of points, primarily noisy outliers.
}
\label{tab:residual_thresholding_eval}
\centering
\begin{tabular}{ p{2.5cm}p{1.1cm}p{1.3cm}p{.8cm}p{0.8cm}}
\toprule
Residual Threshold & \# Points & MedAE (mm) & RMSE (mm) \\
\midrule
No Filter & 809,281 & 1.29 & 6.39 \\
0.10     & 806,152 & 1.28 & 6.18 \\
0.05     & 775,710 & 1.26 & 4.62 \\
0.02     & 621,130 & 1.26 & 1.53 \\
0.01     & 429,923 & 1.25 & 1.52 \\
\bottomrule
\end{tabular}
\end{table}

\section{Pattern Evaluation}
\label{sec:pattern_design}

\begin{figure*}
    \includegraphics[width=\textwidth]{fig/patterns/pattern_evaluation_reconstructions.pdf}
    \caption{{\bf Reconstructions with Different Patterns}. We calibrate 11 different LUTs and reconstruct the same scene: a static pawn. In the top row, we display the patterns used for calibration and reconstruction. In the middle, we display the reconstructed point cloud -- we would like to highlight that the quality of reconstruction with \systemname improves as more channels are included in the LUT. This is because the confusion of a naive lookup decreases as more channels are introduced. In the bottom line, we also report the root mean squared error (RMSE) in millimeters of the reconstructed cloud and the ground truth geometry of the pawn, further validating that, as more channels are introduce, the more precise the reconstruction becomes.}
    \Description[Testing Different Patterns]{We calibrate 11 different lookup tables, ranging from a single channel to eleven channel, and test them on the same scene: a pawn. There are some visible artifacts, such as aliasing and some stair-like errors in the reconstruction at lower pattern count. However, at 11 channels, these artifacts disappear, validating our intuition that more channels reduce error in reconstruction.}
\label{fig:static_patterns}
\end{figure*}

The mechanism of \systemname relies on establishing per-pixel mapping from observed color to depth. Ideally, this mapping is unique with each color corresponding to a unique depth. The extent to which this is true in practice plays a critical role in reconstruction accuracy. \rev{If two or more depth values have similar colors stored in the LUT during calibration, then, during reconstruction, a simple lookup search will often yield incorrect depths. In Fig.~\ref{fig:optimized_patterns}, we plot what the confusion matrix looks like for a {\it Spiral} pattern -- the confusion is computed as the Euclidean distance between a vertical stripe and all other stripes and ranges from 0 (most dissimilar) to 1 (most similar). The perfect, although impossible, pattern would achieve a confusion matrix with 1s in the diagonal and zero elsewhere.}

We evaluated reconstruction performance across a wide range of pattern designs in the static setting. This included \rev{established structured light patterns, such as Gray code and space-filling curve Hilbert patterns, as well as} patterns encoded in 3, 6, and 9 color channels. \rev{For our set-up, we used a $1920\times1080$ resolution projector; therefore, our patterns were designed as $1920$ vertical stripes, each with a unique color sampled from smooth functions (top of Fig.\ref{fig:static_patterns}).}

The results, shown in Table \ref{table:patterns_comparisons}, indicate that increasing the number of channels generally improves reconstruction accuracy. We also observed that different patterns tend to favor different object geometries, reflecting trade-offs between pattern smoothness, spatial coverage, and robustness to ambiguity. \rev{You can find what the Bunny looks like in Figs.~\ref{fig:static_comparisons},~\ref{fig:dynamic_scenes_450fps}, ~\ref{fig:bunny_defocus},~\ref{fig:bunny_linear_motion}; the Irregular Structure in Figs.~\ref{fig:failures},~\ref{fig:static_figures_pages}, and \ref{fig:c2f}; the Dodo in Fig.~\ref{fig:c2f}; and the House in Fig.~\ref{fig:static_lights_on}.

We additionally visually display the results with the same patterns from Table \ref{table:patterns_comparisons} in Fig.~\ref{fig:static_patterns} with the Pawn. We also see visually that increasing the number of channels generally improves reconstruction accuracy. We would also like to highlight some of the artifacts of reconstruction: for instance, the wiggles in 6-channel {\it Hilbert} and the aliasing in 3-channel {\it Spiral}. These artifacts are due to strong confusion in \systemname and mostly arise from sensor noise during both calibration and scanning -- if the pattern is already designed with similar color values, then even small amounts of sensor noise can cause different depth values to end up with extremely similar colors stored into the LUT. With 11 channels (and no denoising of the LUT), the confusion of \systemname decreases and the artifacts disappear in the pawn (right-most reconstruction of Fig.~\ref{fig:static_patterns}).}

Given that our method is designed to generalize on both patterns and object geometries, and that the dynamic setting requires a single RGB pattern (i.e., 3 channels), we choose the spiral pattern for high-speed reconstruction. While Hilbert patterns yielded the lowest errors in static scans, they degrade sharply when blurred, which is an issue in high-speed capture with shallow depth of field and film-based projection. Spiral patterns, by contrast, maintain non-ambiguity more reliably under blur and warping (see Fig.~\ref{fig:optimized_patterns}), making them a better fit for dynamic scenes. 

Future work may explore tailored pattern design and optimization for specific hardware and scene conditions, though our current results already demonstrate that simple, unoptimized patterns can yield accurate reconstructions across a range of setups.


\begin{figure}
    \centering
    \includegraphics[
    width=\columnwidth,
    trim=200pt 1980pt 1200pt 0pt, clip
]{fig/patterns/optimized_patterns.pdf}
\includegraphics[
    width=\columnwidth,
    trim=200pt 50pt 1200pt 2700pt, clip
]{fig/patterns/optimized_patterns.pdf}
    \caption{{\bf Spiral Pattern.} Representative {\it Spiral} pattern (left) with low probability of confusions between different image column pairs (middle). The representation of the pattern as a path in an RGB color cube is shown on the right. The linear component of the pattern in the green channel is particularly useful for identifying when a pixel is out of bounds of the reconstructible volume.}
\Description[Spiral Pattern]{We display the spiral pattern, which we use in our dynamic scenes. This pattern is a straight line in the green channel, with the red and blue channels oscillating like sinusoids with modulated amplitudes.}
    \label{fig:optimized_patterns}
\end{figure}


\begin{table*}[t]
\centering
\caption{Comparison of performance of our proposed \systemname method with several different patterns in static setup. We report the mean absolute error (MAE), median absolute error (MedAE), and root mean squared error (RMSE), all in millimeters, for each reconstructed point cloud with respect to its ground-truth geometry. For a visualization of what the patterns look like, refer to Fig.~\ref{fig:static_patterns}. For a visualization of the objects, check Fig.~\ref{fig:static_comparisons} for the Bunny, Fig.~\ref{fig:c2f} for the 3D printed Dodo, Fig.~\ref{fig:static_figures_pages}, and Fig.~\ref{fig:static_lights_on} for the House. Hilbert and Spiral patterns stand out as better when constrained to 3 channels.}
\label{table:patterns_comparisons}
\begin{tabular}{ p{1.7cm}p{1.5cm}p{0.7cm}p{0.7cm}p{0.7cm}p{0.7cm}p{0.7cm}p{0.7cm}p{0.7cm}p{0.7cm}p{0.7cm}p{0.7cm}p{0.7cm}p{0.7cm}}
 \toprule
 & & \multicolumn{3}{c}{Bunny} & \multicolumn{3}{c}{Irregular Structure} & \multicolumn{3}{c}{Dodo} & \multicolumn{3}{c}{House} \\
  \cmidrule(lr){3-5} \cmidrule(lr){6-8} \cmidrule(lr){9-11} \cmidrule(lr){12-14} 
 Pattern & \# Channels &MAE &MedAE &RMSE &MAE &MedAE &RMSE &MAE &MedAE &RMSE &MAE &MedAE &RMSE\\
 \midrule
 Ramp &1
 &4.54&3.62&5.83   
 &4.03&3.12&5.51
 &7.99&6.11&12.10
 &5.48&4.09&7.96\\
 Ramp+Sine &2 
 &4.22&3.23&5.66
 &1.39&1.66&5.05
 &10.76&5.16&16.61
 &4.82&3.04&7.75\\
 Spiral &3 
 &5.21&0.46&10.42   
 &2.96&1.26&6.61
 &17.40&0.95&31.07
 &41.73&4.48&70.4\\
 Stairs &3
 &1.05&0.54&1.89   
 &1.53&1.26&1.96
 &5.61&0.81&13.36
 &2.81&1.78&6.47\\
 Random &3
 &3.90&0.49&11.82   
 &1.97&1.17&5.87
 &3.27&0.65&12.16
 &8.72&1.82&25.24\\
 Hilbert &3
 &0.86&0.49&1.46   
 &1.39&1.21&13.50
 &2.63&0.80&7.82
 &2.21&1.71&2.77\\
 Spiral+Hilbert &6
 &0.69&0.39&1.12   
 &1.25&1.20&1.39
 &1.06&0.58&4.75
 &2.28&1.61&4.65\\
 Hilbert &6 
 &0.92&0.46&1.89   
 &1.27&1.22&1.36
 &0.82&0.55&1.84
 &2.08&1.68&2.77\\
 Colors&9 
 &0.68&0.39&1.92   
 &1.24&1.20&1.35
 &0.57&0.54&0.89
 &1.87&1.58&2.45\\
 Spiral+Hilbert &9
 &0.60&0.40&1.03   
 &1.25&1.20&1.34
 &0.55&0.52&0.66
 &1.86&1.57&2.57\\
 Gray Code &11
 &0.50&0.32&1.26   
 &1.28&1.24&1.36
 &0.50&0.48&0.74
 &1.86&1.57&2.46\\
 \bottomrule
\end{tabular}
\end{table*}

\rev{\section{Film Development for Patterns}
\label{sec:film_development}

\begin{figure}
\centering
\includegraphics[width=0.86\columnwidth]{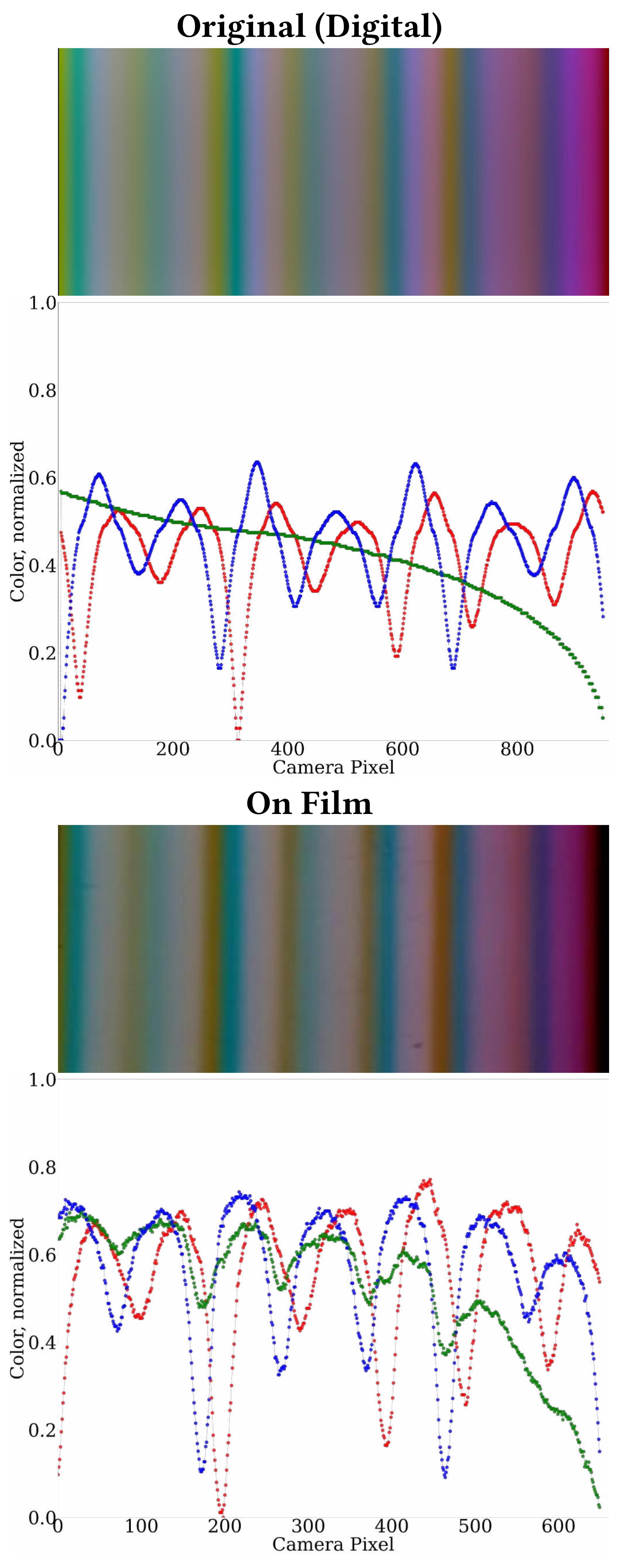}
\caption{{\bf Film Development.} The {\it Spiral} pattern is designed as a monotonic slope in green and two shifted sinusoidal functions with parametrized amplitudes in the red and blue channels. When the film is developed with the {\it Spiral} pattern on it, the film granularity and the color mixing impact the quality of the pattern. This introduces confusion to the lookup tables in our dynamic setting, which we try to overcome with strategies such as coarse-to-fine reconstructions and enforcing temporal consistency. 
}
\Description[Pattern on Film]{The original, digital spiral pattern, when developed into film, has a lot of artifacts that are currently hard to model. Mainly, we visualize that there is intense color mixing and a grainy texture from the film.}
\label{fig:film_development}
\end{figure}

Our analog projector can flicker high-power LEDs at extremely high rates. In order to get a three-channel pattern projected onto the scene, we need to place a transmissive sheet in front of the projector LED. For the prototype presented in this paper, we opted for a 35mm film.

We use a Canon EOS-1N analog camera with Tamron 90mm f/2.8 lens and use both FUJIFILM Reverse Film Fujichrome Velvia 50 135 Film and Kodak Ektar 100 Professional ISO 100. We design patterns as 1920 vertical stripes and use the same DLP projector and shine these various patterns onto a flat white screen. We place the camera atop the projector and expose each pattern on film at 1/6, 1/10, 1/12, 1/15, 1/20, 1/30 second. We use a commercial third-party to develop the film.

The patterns are designed such that every vertical stripe is its unique color and with enough color distinction between different depths to avoid noise. Developing the film introduces color mixing that is hard to model and provokes confusion in the lookup table (see Fig.\ref{fig:film_development} for the original pattern design on the left and the developed film on the right). The color mixing is also heavily dependent on the batch of film roll and the Bayer filter employed by the high-speed camera, making it even harder to predict beforehand any artifact that comes from exposing and developing the film. 

In addition to color mixing, the film, after developing, obtains a grainy texture, which further causes issues in the lookup table. Most of the patterns we use are designed as smooth functions along the RGB cube, so the granularity introduced by the film goes against the smoothness. To overcome that, we purposefully keep the analog projector out of focus for the whole volume of reconstruction; by projecting a defocused pattern, we can achieve a smoother ray for the lookup table and avoid the granularity of the film. 

Due to the difficulties of modeling color mixing and other random effects of developing film, our data capture currently relies on exposing several patterns on film and pick the best one based on our experiments. There are likely better ways to produce the film and alternatives to film itself (such as stained glass), which we leave as future work.}

\section{\systemname Implementation Details}
\label{sec:lookup_details}

Reconstruction with \systemname, in its unoptimized form, is a single lookup on each observed pixel $I_i$ against a depth-to-color mapping function $C_i(d)$:
\begin{equation}
d_i^* = \argmin_{d \in \mathcal{D}} ||C_i(d) - I_i||_2,
\end{equation}
where the search is done on a discrete set of points $\mathcal{D}$.

This naive implementation works well in our static setting, where  noise is minimal since we use a high-end monochromatic camera and a high-end projector. However, bringing the naive implementation of the pipeline to our dynamic setting is quite challenging. In this section, we describe the steps we found necessary to achieve high-quality reconstructions in adversarial conditions.

\subsection{Denoising.}
\label{subsec:denoising} Our lookup tables are a collection of densely sampled color values that are a $H \times W \times n \times 3$ tensor, where $H$ and $W$ are the height and width of the image, $n$ is the number of depth steps. The patterns we use are often smooth in comparison to binary patterns, i.e. they do not change rapidly. This is particularly important in compressing and denoising our data. There is a plethora of strategies to denoise data; \rev{here, we explore a few of such strategies and discuss their respective advantages and disadvantages.}

\begin{figure}
    \centering
    \includegraphics[width=0.99\columnwidth]{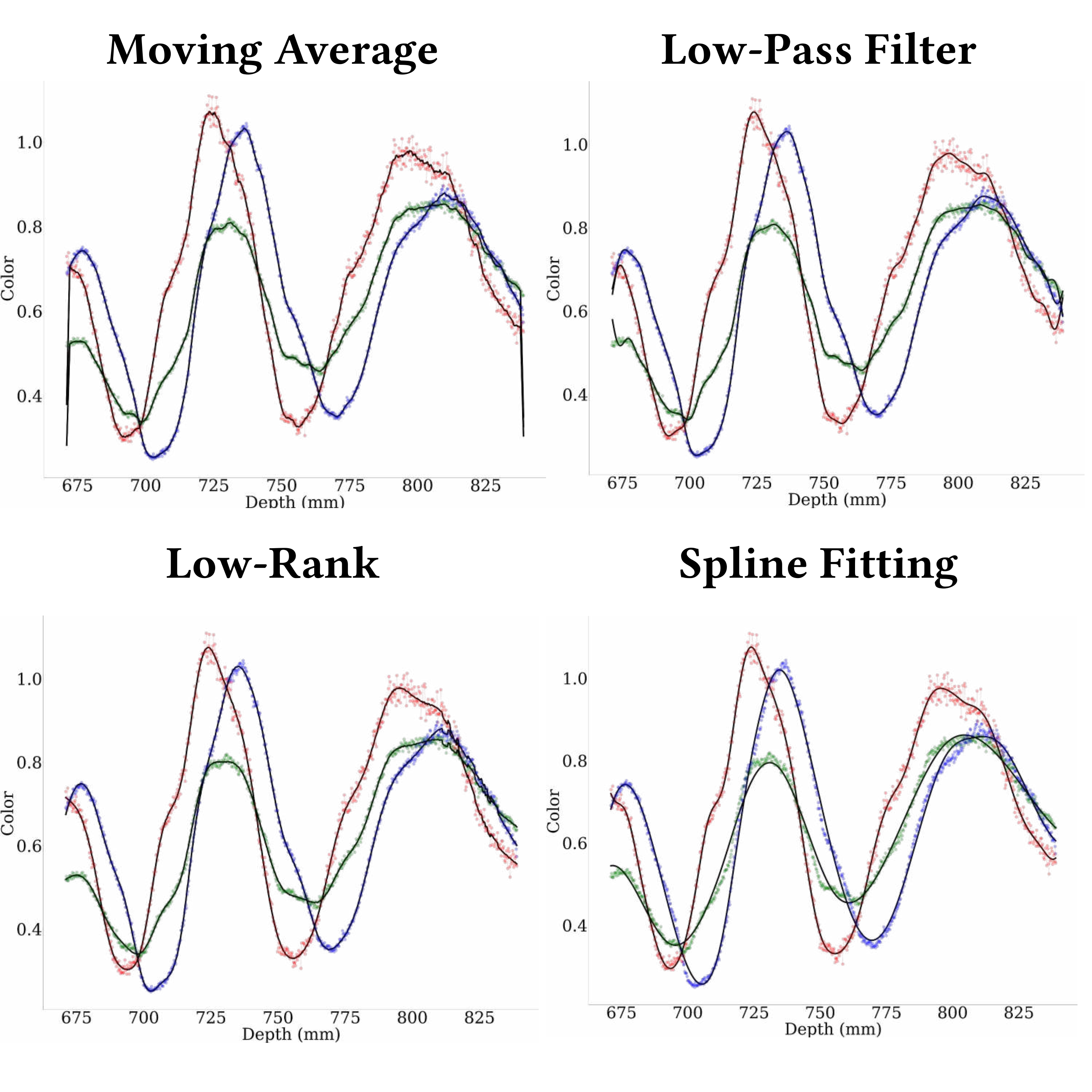}
    \caption{{\bf Example of Denoised Ray.} We illustrate, with color, what the red, green, and blue channels of a single ray looks like. We overlay in black the denoised rays from: a moving average of size 11 (top left), a low-pass filter with a cutoff at frequency 25 (top right), a low-rank approximation using the top 25 singular values (bottom left), a cubic B-Spline with smoothing parameter of 0.1 (bottom right). The black curves follow the trend of the red, green, and blue curves well, while neglecting some noisy, high-frequency components.}
    \Description[Denoising the Same Ray]{We try four different methods to remove noise from the red, green, and blue channels of our spiral lookup table. We plot the same ray with moving average, a low-pass filter, a low-rank approximation, and spline fitting. They all are successful in removing noise and capturing well the main structure of the ray.}
    \label{fig:denoising_rays}
\end{figure}

\rev{\paragraph{Moving Average.}
The moving average is the simplest (and fastest) form of reducing variance in a function. In a single ray, we use a fixed window size to average the color intensity at a certain depth point with its neighboring depths (top left of Fig.~\ref{fig:denoising_rays}). This does nothing to reduce storage and it maintains the original depth discretization from calibration.}

\rev{\paragraph{Low-Pass Filter.}
We denoise each channel of the LUT independently. We apply a Fourier Transform to each ray of every channel, bringing them into the frequency domain. We then apply a low-pass filter and perform the inverse Fourier Transform to bring the ray back into the spatial domain, but now with the high-frequency details removed. With Fast Fourier Transform algorithms implemented in {\it numpy} and {\it scipy}, this is a fast way to get rid of noise in the data, although it maintains the depth discretization from calibration and does nothing to reduce storage requirements.}

\paragraph{Low-Rank Approximation.}
We perform the low-rank approximation independently on each channel of the LUT, e.g., the  $H \times W \times n$ tensor representing the red channel. 
We gather all pixels' values into a matrix $\AB \in \mathbb{R}^{k \times n}$ where $k = HW$ gathers all the pixels. We then construct a low-rank approximation by taking the SVD $\AB = \UB \SigmaB \VB^\top$ where $\UB, \VB$ are orthonormal matrices and $\SigmaB$ has the singular values $\sigma_0, \ldots, \sigma_{n-1}$ along its diagonal.

By truncating the singular values to the top $r$ and reconstructing the matrix $\AB_r = \UB_r \SigmaB_r \VB^{\top}_r$, one obtains the closest (in the Frobenius norm sense) rank-r matrix~\cite{eckart1936approximation}. The accrued error is given by the  Frobenius norm of the difference
\begin{align*}
    \|\mathbf{A} - \mathbf{A}_r\|_F^2 = \sum_{i=r}^{n-1} \sigma_i^2, 
\end{align*}
and so by choosing a value of $r$ for which most of the discarded singular values are much smaller than the top $r$ singular values is desirable to maintain accuracy while denoising the data and compressing the matrix for storage.

\begin{figure}
    \centering
    \includegraphics[width=.8\columnwidth]{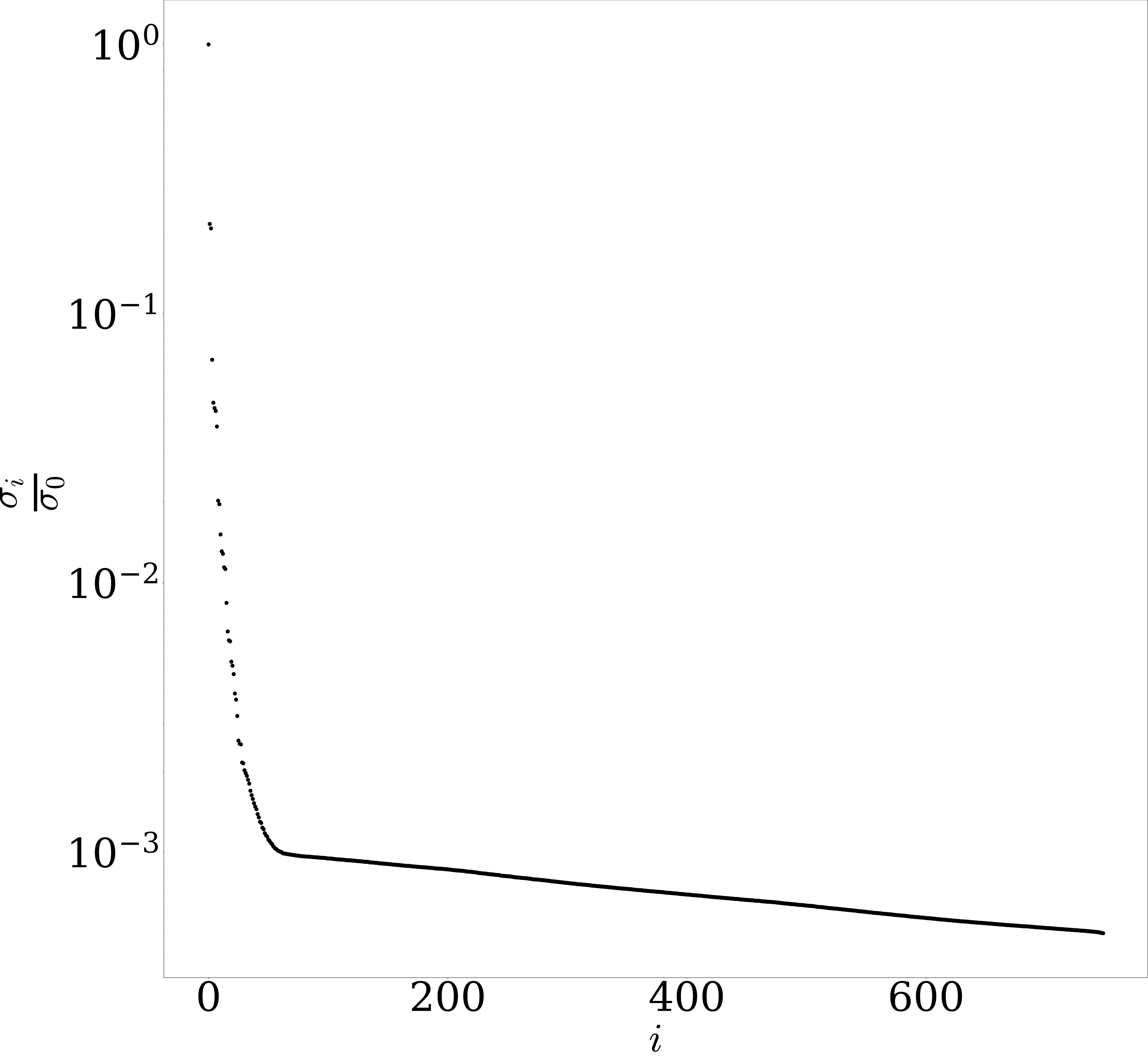}
    \caption{{\bf Singular Values of Red Channel of lookup table.} We plot all 750 singular values normalized by the top singular value $\sigma_0$. The mass of the red channel is concentrated in the top 50 singular values, meaning we can safely truncate the vast majority of bottom singular values while maintaining the overall structure of the matrix.}
    \label{fig:singular_values}
    \Description[Singular Values of Red Channel]{We plot the 750 top singular values, normalized by the highest. We visualize that there is a sharp decline in the mass of the singular values, with most being concentrated in the top 50. After that, the mass remains extremely small with a small downward slope.}
\end{figure}

In Fig.~\ref{fig:singular_values}, we plot all 750 singular values of the red channel of the lookup table we use for our dynamic scenes. The plot shows that the mass of the matrix is concentrated in the top 50 singular values, which means that we can safely truncate the vast majority of these singular values while maintaining accuracy. In the bottom left corner of Fig.~\ref{fig:denoising_rays}, we plot the example of a ray from the same lookup table, after keeping only the top 25 singular values for all color channels -- the low-rank approximation accurately captures the trend in the color intensity through depth, while discarding most of the high frequency (and thus likely noise) components of the ray.

\rev{Low-rank approximation is a powerful and fast tool to reduce storage requirements of the lookup tables: in the static set-up, the 3-channel {\it Spiral} pattern with 700 depth steps and 3.7~Megapixel resolution takes $38.8$\,GB to store; after low-rank approximation with the top 35 singular values for all channels, that storage is reduced to $1.8$\,GB, 21 times smaller than the original one. The depth discretization, however, remains the same.}

\rev{\paragraph{Spline Fitting}
We use, per-ray, cubic B-splines, creating a smooth depth-to-color mapping; moving average, low-pass filter, and low-rank approximation all maintain the original discretization from \systemname calibration. With splines, instead, we have a direct way to interpolated data and further reduce error of reconstructed clouds. In our static setting with the 9-channel {\it Spiral+Random+Stairs} pattern, RMSE of the pawn decreases from 1.44mm to 0.70mm (Fig.~\ref{fig:pawn_spline}).

\begin{figure}
    \centering
    \includegraphics[width=0.95\columnwidth]{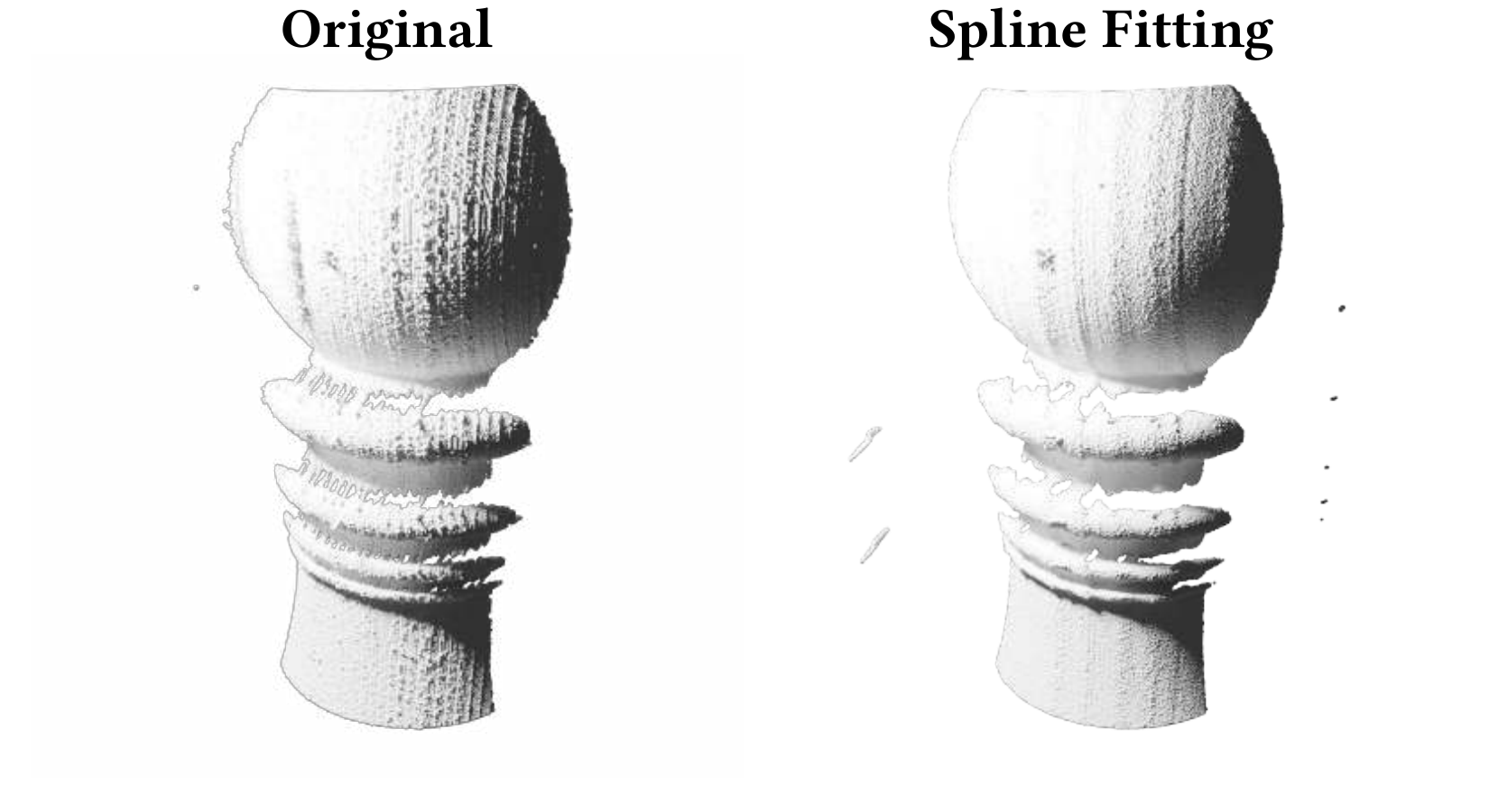}
    \caption{{\bf Spline Fitting with 9-channel pattern.} We denoise the LUT by fitting cubic splines per channel per ray, setting a smoothing parameter of 0.1. The splines, in addition to denoising the data, allow for a smooth depth-to-color mapping, further reducing the error in reconstruction.}
    \Description[Reconstructing Pawn with Spline]{We display the pawn reconstructed with no denoising and with spline fitting. Side-by-side, the spline fitting shows how it removes a lot of the errors in reconstructing the pawn.}
    \label{fig:pawn_spline}
\end{figure}

Spline fitting also reduces the storage requirements for the lookup tables, but not as dastrically as the low-rank approximation. In the static set-up, the 3-channel {\it Spiral} pattern with 700 depth steps and 3.7~Megapixel resolution takes $38.8$\,GB to store; with cubic B-spline fitting that storage is reduced to $5.4$\,GB, 7 times smaller than the original one but 3 times larger than low-rank.}

\subsection{Coarse to Fine (C2F).}
\label{subsec:c2f} We first downsample our captured image to very few pixels. We search, for this subsample of pixels, the whole range of depth values and pick the minimum. Then, when double the resolution and choose limit the subsample to be half of the whole depth range. However, we still ensure that the residual in this shortened range is below a threshold $\epsilon$ -- otherwise, we double the range back up until we search all depth points in the LUT or we find a depth position below threshold $\epsilon$.

\rev{C2F effectively reconstructs, first, a coarse geometry of the object and uses it as a prior for higher-resolution reconstructions. C2F forces the reconstructed points to lie on the main structure of the object. In Fig.~\ref{fig:c2f}, we reconstruct the pawn, a 3D printed dodo, and the precisely-machined irregular structured with the 3-channel {\it Spiral} pattern. On the left, the two objects are reconstructed with the naive method; on the right, we use C2F in the reconstruction pipeline. For the pawn, the RMSE drops from $6.66$\,mm (993,296 points) to $0.76$\,mm (947,678 points); for the dodo, the median absolute error was $0.95$\,mm and the RMSE was $31.07$\,mm (1,169,540 points), but with C2F they drop to $0.61$\,mm and $6.26$\,mm (869,780 points) respectively; for the irregular structure, the RMSE drops from $6.61$\,mm (2,451,315 points) to $2.36$\,mm (2,153,456 points).

\begin{figure}
    \centering
    \includegraphics[width=0.95\columnwidth]{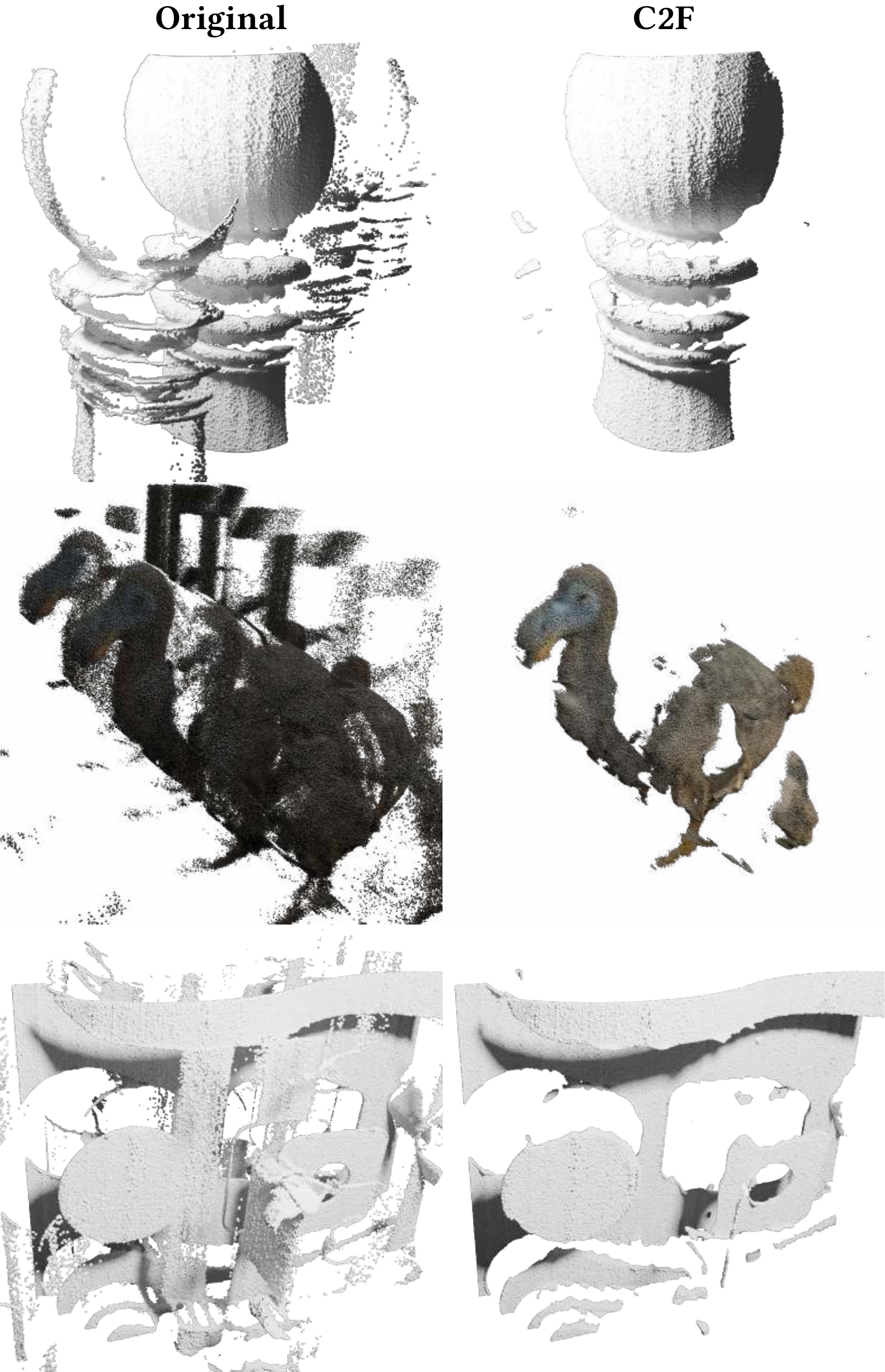}
    \caption{{\bf Coarse-to-Fine (C2F) with 3-channel pattern.} We first use a lower-resolution normalized image to reconstruct a coarse geometry of the object. Then, we increase the resolution of the image but limit the range of depths to search in the lookup table, effectively using the coarser geometry as a prior for the full-resolution reconstruction. In the original reconstruction, the pawn has many points wrongly reconstructed; with C2F, several of these points get pushed back into the correct main structure.}
    \Description[Reconstructing with Coarse-to-Fine approacg]{We display the pawn, a 3D printed dodo, and an irregular structure, reconstructed with the naive method and with coarse-to-fine approach. Side-by-side, the coarse-to-fine approach shows how it forces a lot of the incorrectly reconstructed points to be pushed back into the main structure. Coarse-to-fine effectively leverages neighborhood information to restrict the dept range. Even with coarse-to-fine, there are still regions incorrectly reconstructed.}
    \label{fig:c2f}
\end{figure}}

\subsection{Temporal Consistency.}
\label{subsec:tc} Given the reconstruction of the previous frame, we avoid searching through the whole depth range again and instead search around a small parameter $\delta$ millimeters around the previous depth and find the minimum. If the residual in this shortened range is above a threshold $\epsilon$, we double the search to $2\delta$. We repeat until we search all depth points in the LUT or we find a depth position below threshold $\epsilon$. The parameters $\delta$ and $\epsilon$ can be modified; a smaller $\delta$ can lead to artifacts, where; a smaller $\epsilon$ can lead to longer reconstructions as most pixels will end up searching the whole ray to find the minimum. We set $\delta = 5$\,mm in all our reconstructions and $\epsilon$ to 0.1 or 0.2, depending on how noisy the results were.

\begin{figure}
    \centering
    \includegraphics[width=\columnwidth]{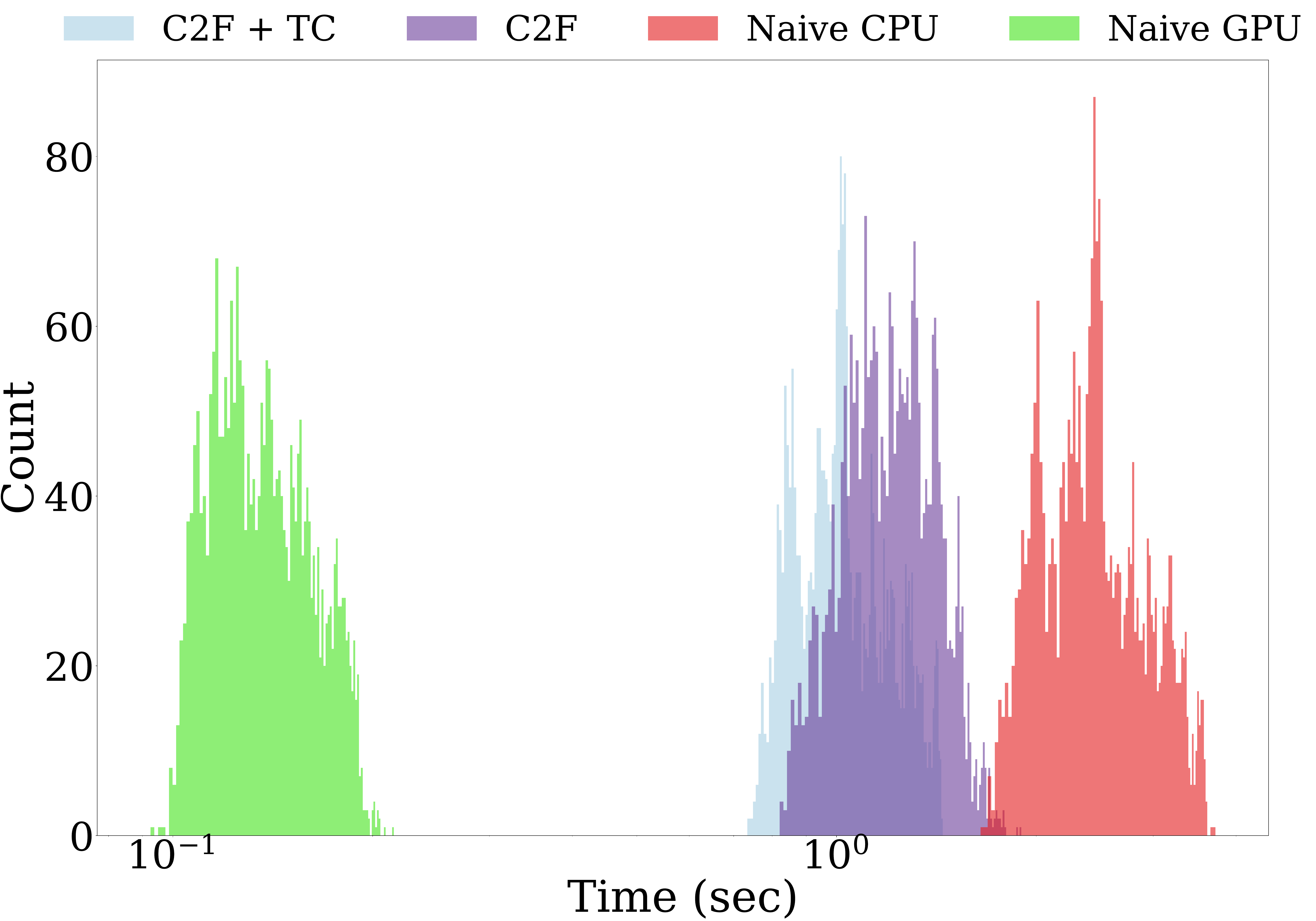}
    \caption{{\bf Times for Depth Decoding using Different Strategies.} We collect the per-frame time to decode depth for ten different dynamic scenes. We use the naive (with low-rank approximation) approach that is implemented for the static reconstructions; we also use the coarse-to-fine (C2F) approach, and, ultimately, add the temporal consistency (TC), which gives the shortest median time for depth reconstruction. Naive CPU, C2F, and TC reconstructions are done using one core of an 
    AMD EPYC 9654P processor
    clocked at 3.7 GHz. The naive GPU reconstruction is done in an NVIDIA RTX A6000 GPU.}
    \Description[Plot with Decoding Times]{A histogram plot shows the decoding times with different methods: naive on the CPU, coarse-to-fine, coarse-to-fine with temporal consistency, and naive on the GPU. The slowest one is naive on the CPU, concentrating around 2 seconds per frame. The fastest is naive on the GPU, concentrating around 0.1 second per frame. Both the coarse-to-fine and coarse-to-fine with temporal consistency concentrate around 1 second.}
    \label{fig:times}
\end{figure}

Finally, we show that both C2F and TC strategies, in addition to improving the reconstruction quality in the dynamic setting, also reduce (by at least half) the average time of decoding depth for a full frame (Fig.~\ref{fig:times}). All reconstructions are done using 1 core of an AMD EPYC 9654P processor clocked at 3.7 GHz. \rev{We also implement, with PyTorch, an argmin search in the GPU, using an NVIDIA RTX A6000 GPU.} \looseness=-1

\rev{\subsection{Multilayer Perceptron}
All the results presented in this paper are reconstructed by a per-pixel lookup against a depth-to-color mapping $C_i(d)$:
\begin{equation}
d_i^*=\arg\min_{d\in\mathcal{D}} \left\lVert C_i(d)-I_i \right\rVert_2,
\end{equation}
where the search is performed over a discrete set $\mathcal{D}$.

As an unoptimized proof of concept prototype, we also learn the lookup search with a multilayer perceptron (MLP) for depth decoding. We use five input features per ray $(i/H,\ j/W,\ r,\ g,\ b)$, where $H$, $W$ are the height and width of the camera resolution, $i,\ j$ are the row and column of a certain pixel, and $r,\ g,\ b$ are the normalized color values associated with the pixel. We apply a positional encoding with $L{=}10$ frequency levels, yielding a 105-dimensional input. The network is a compact MLP with two hidden layers of widths $(256,\,128)$ and a final $1$-unit output; the activation is a {\it Swish} function (where Swish$(x) = \frac{x}{1-e^{-x}}$). We train with AdamW (learning rate $10^{-3}$) for up to 100 epochs (validation RMSE typically plateaus by around 20 epochs; we report the best epoch) using mean squared error between predicted and stored depth. Depth is normalized to $[0,1]$ during training and mapped back to mm for reporting.

Training data are built from a $300{\times}300$ crop of the LUT (masking depths in $[500, 2000]$\,mm), giving a total of $67{,}409{,}986$ rays. Each epoch processes $\approx 4{,}115$ mini-batches of up to $16{,}384$ rays (shuffle; last batch is smaller). On a single NVIDIA H100~(80\,GB HBM3), each epoch takes $\sim$2--3 minutes; over $100$ epochs this corresponds to roughly $3$--$5$ hours wall-clock in our environment. The model has $60{,}161$ parameters ($\approx 0.23$\,MB in single floating-point precision). Inference-time reconstruction runs in about $1$--$2$ seconds per frame on the same GPU.

\begin{figure}
  \centering
  \includegraphics[width=0.95\columnwidth]{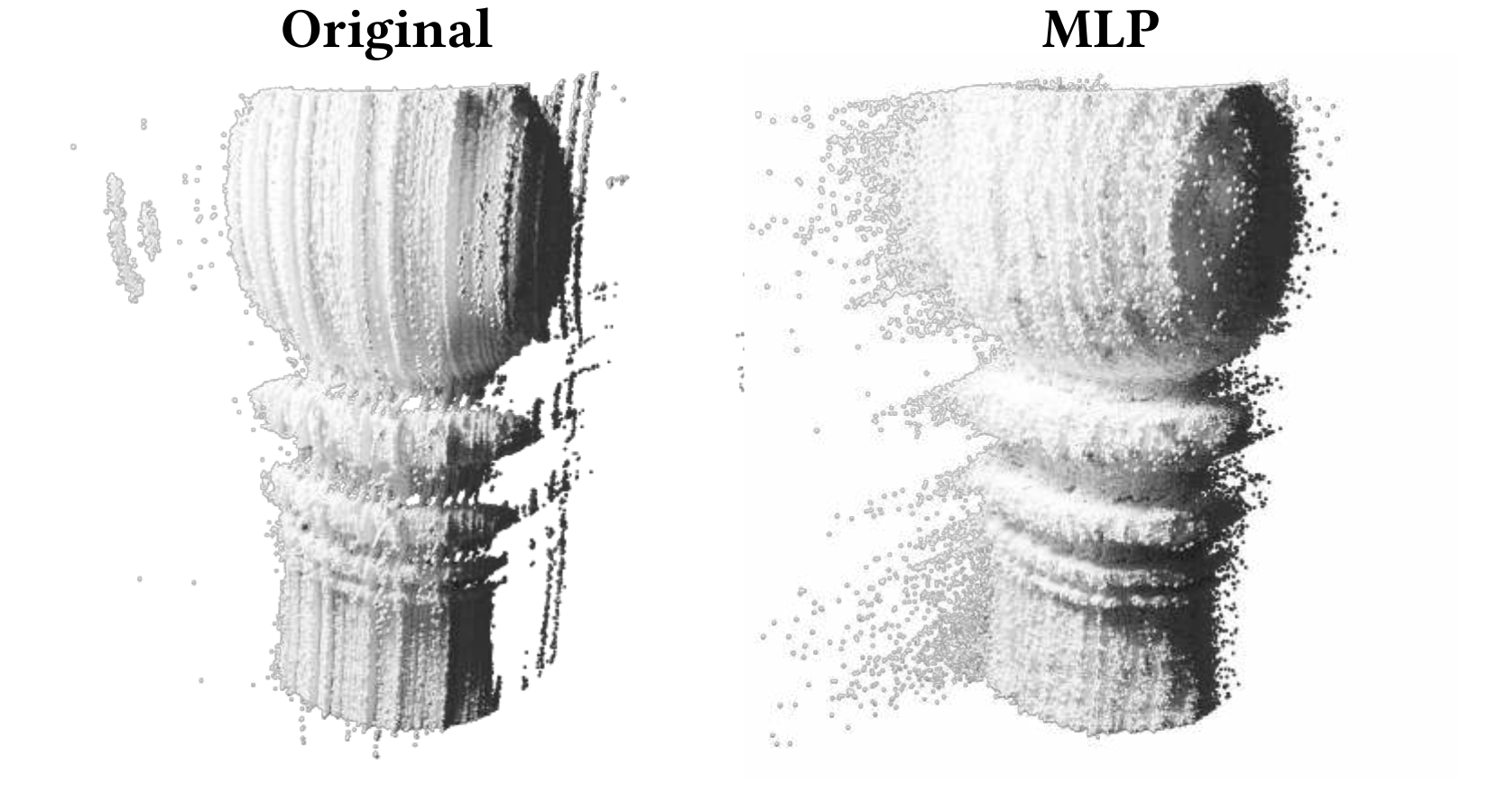}
  \caption{\textbf{MLP depth prediction with a 3-channel pattern.} We train an MLP on the 3-channel {\it Hilbert} pattern and predict the pawn geometry.}
  \label{fig:pawn_mlp}
  \Description[Reconstructing Pawn with Multilayer Perceptron]{We display the pawn reconstructed with our naive lookup approach and with a multilayer perceptron. Side-by-side, they both show artifacts in the reconstruction, but the multilayer perceptron shows potential in incorporating deep learning into our 3D scanning set-up.}
\end{figure}

Fig.~\ref{fig:pawn_mlp} shows a side-by-side comparison using the 3-channel {\it Hilbert} pattern: direct lookup (994{,}319 points, RMSE $1.86$\,mm) vs.\ the trained MLP (991{,}706 points, RMSE $1.03$\,mm).
}

\section{Additional Dynamic Reconstructions}
\label{sec:additional_dynamic}

In this section, we show three dynamic sequences reconstructed with our high-speed prototype and \systemname.

We placed the Stanford bunny on a platform and attached it to our linear stage. We begin capturing and set the linear stage to move $31.25$ millimeters at 12.50\,mm/s. We reconstruct its linear motion at 450~fps (Fig.~\ref{fig:bunny_linear_motion}). We use these reconstructions to validate both the accuracy of the reconstructed point cloud, achieving a root mean square error of 3.27\,mm. This result is mentioned in our quantitative analysis of the high-speed scanning regime (Section~\ref{sec:quantitative_analysis}).

\begin{figure*}[t]
    \centering
    \includegraphics[width=\columnwidth]{fig/dynamic/bunny_31dot25_mm_motion.pdf}
    \caption{{\bf Bunny in Linear Motion.} We reconstructed, at 450~fps, the bunny being pushed by a linear stage. This image displays two point clouds together in the same coordinate system -- the one to the left (orange) is the bunny at the beginning of the motion, before the linear stage had started moving; the one to the right (blue) is the bunny at the end of the motion, 1249 frames later, just after the linear stage had stopped moving. We set the speed to 12.50\,mm/s and the distance to 31.25\,mm -- we recovered a speed of 12.56\,mm/s and a distance of 31.52\,mm with our reconstructions.}
    \Description[Additional Reconstruction: Bunny on Linear Stage]{There are four views of the bunny before moving and the bunny after moving. Even with a small motion of 31.25 millimeters, our method reconstructs the linear motion accurately.}
    \label{fig:bunny_linear_motion}
\end{figure*}

We grab a small, white, flat plane (approx 20cm $\times$ 15cm) and move (rotate and translate) it in the volume of reconstruction. We reconstruct around six seconds of motion at 450~fps, resulting in over 3,000 frames and more than one billion reconstructed points (Fig.~\ref{fig:dynamic_flat_plane}). We assume that the plane is and remains perfectly flat throughout the motion; this is an additional way to quantitatively evaluate the performance of \systemname and our prototype at high frame rates for each frame. We fit a perfect plane to our reconstruct point clouds and measure the total deviation from this assumption: we obtain a root mean square error to be 3.83\,mm, validating precision of our system at high frame rates. 

\begin{figure*}
    \centering
    \includegraphics[width=\linewidth]{fig/dynamic/small_white_plane_moving.pdf}
    \caption{{\bf Flat Plane in Motion.} We reconstructed, at 450~fps, a flat plane moving around the reconstruction volume. The average root mean square error was less than 4\,mm, validating precision of \systemname and our prototype at high frame rates.}
    \Description[Additional Reconstruction: Flat Plane]{There are six frames, each 1.1 second apart, of a flat plane being moved and rotated. Our method captures the flatness of the plane and its motion well.}
    \label{fig:dynamic_flat_plane}
\end{figure*}

We grab a sheet of paper and tear it in half. We reconstruct around three seconds of it at 100~fps (Fig.~\ref{fig:ripping_paper}).

\begin{figure*}
    \centering
    \includegraphics[width=\linewidth]{fig/dynamic/ripping_paper.pdf}
    \caption{{\bf Ripping Paper.} We reconstructed, at 100~fps, a sheet of paper being split in half.}
    \Description[Additional Reconstruction: Ripping Paper]{There are six frames, each 0.5 second apart, of a piece of paper being ripped in half. Our method captures the motion well.}
    \label{fig:ripping_paper}
\end{figure*}

We also attempted to use our high-speed prototype with \systemname to reconstruct a balloon popping (Fig.~\ref{fig:balloon_popping}). Unfortunately, even at 1450~fps, the motion of the balloon as it explodes is too fast for our scanner. Since we take two images (a pattern projected and a white flash), if these two images do not overlap much, \systemname will most likely fail to recover any relevant depth.

\begin{figure*}
    \centering
    \includegraphics[width=\linewidth]{fig/failures/big_balloon_popping.pdf}
    \caption{{\bf Failure of Balloon Popping.} We set our camera exposure to 0.3 millisecond and recording frame rate at 2,900~fps. The motion of the balloon popping, however, is too fast -- consecutive frames do not overlap much. Since \systemname normalizes a pattern image against a white flash, the normalized image cannot be computed with very different pattern and white flash.}
    \Description[Failure when motion is too fast]{There are six images, each 0.335 milliseconds apart, of an air balloon popping. Even with an extremely high frame rate, the balloon moves too fast between consecutive frames.}
    \label{fig:balloon_popping}
\end{figure*}

\renewcommand{\theTotPages}{11}

\end{document}